\documentclass{article}

\usepackage[main, final]{neurips_2026}

\usepackage{amsmath}
\usepackage[utf8]{inputenc} 
\usepackage[T1]{fontenc}    
\usepackage{hyperref}       
\usepackage{url}            
\usepackage{booktabs}       
\usepackage{amsfonts}       
\usepackage{nicefrac}       
\usepackage{microtype}      
\usepackage{xcolor}         
\usepackage{graphicx}

\definecolor{best_color}{RGB}{200,120,0}      
\definecolor{second_color}{RGB}{80,125,190}   
\definecolor{third_color}{RGB}{180,158,214}

\newcommand{\best}[1]{\textbf{\textcolor{best_color}{#1}}}
\newcommand{\second}[1]{\textbf{\underline{\textcolor{second_color}{#1}}}}
\newcommand{\third}[1]{\textcolor{black}{#1}}

\newcommand{\bX}{\boldsymbol{X}}
\newcommand{\bK}{\boldsymbol{K}}

\usepackage{amsthm}
\newtheorem{theorem}{Theorem}[section]
\newtheorem{proposition}[theorem]{Proposition}

\title{Vendi Anomaly Scores for Efficient and Accurate Anomaly Detection}

\author{%
\begin{tabular}{c}
    \textbf{Amey P. Pasarkar}$^{*,\ddagger}$
    \qquad
    \textbf{Adji Bousso Dieng}$^{\dagger,\ddagger}$ \\[0.6em]
    {\normalfont $^{*}$ Lewis-Sigler Institute for Integrative Genomics, Princeton University} \\
    {\normalfont $^{\dagger}$ Department of Computer Science, Princeton University} \\
    {\normalfont $^{\ddagger}$ \href{https://vertaix.princeton.edu/}{Vertaix}}
\end{tabular}
}

\begin{document}

\maketitle

\begin{abstract}
    Anomaly detection (AD) is a critical for trustworthy machine learning and scientific discovery. Existing methods typically define anomalies through local density, isolation heuristics, or distance from a fitted distribution, making them sensitive to neighborhood scale, partitioning choices, or restrictive distributional assumptions. In this work, we introduce a new paradigm by formulating anomaly detection from a diversity perspective. We propose the \emph{Vendi Anomaly Score} (VAS), which detects anomalies by quantifying how much a sample changes the diversity of the dataset, as measured by the Vendi score, when removed. VAS captures both local redundancy and global data structure without relying on density estimation or partitioning heuristics. Moreover, VAS requires no hyperparameter tuning, instead adapting automatically to the spectral structure of the data. VAS is non-parametric and scales linearly with dataset size. Across 148 benchmark embedding datasets derived from 10 source datasets and 4 embedding architectures, VAS achieves state-of-the-art AD performance. We further validate VAS on large-scale ImageNet experiments, where it remains robust as both contamination rate and dataset scale increase. Code to reproduce our results is openly available at \url{https://github.com/vertaix/vendi-anomaly-scores}.
\end{abstract}

\section{Introduction}
Anomaly detection (AD) is essential for analyzing and curating machine learning datasets and for discovery~\citep{chandola2009anomaly, campolongo2025building}. In particular, by identifying samples that are corrupted, mislabeled, out-of-distribution, or otherwise irrelevant, AD helps maintain data quality and improve model reliability \citep{ahmed2017anomaly,nassif2021machine}. Since modern datasets are increasingly inspected through learned embeddings that encode semantic relationships, effective data curation requires AD methods that operate directly in these embedding spaces \citep{pang2021deep,reiss2021panda,cao2025tad}.

AD algorithms are commonly categorized by the supervision available during training. Supervised methods fit detectors using labeled mixtures of normal and anomalous samples \citep{ma2016supervised,vargaftik2021rade}, while semi-supervised methods assume access to clean data and identify deviations from this reference set \citep{dai2024unsupervised}. Unsupervised AD instead fits directly to the dataset being analyzed, making it especially attractive for modern data curation. As datasets increasingly span diverse modalities, it is often impractical to collect data on anomalies or to construct clean reference sets for every domain, making unsupervised detectors appealing. 

Unsupervised AD methods typically define anomalies through isolation \citep{liu2008isolation}, local density \citep{breunig2000lof}, or distance from a fitted distribution \citep{rippel2021modeling}. Isolation-based methods score points by how easily they can be separated from the rest of the data, with classical examples including isolation forests (iForest) \citep{liu2008isolation}, extended isolation forests (EIF) \citep{hariri2019extended}, and robust random cut forests (RRCF) \citep{guha2016robust}, as well as geometric variants such as the hypersphere ensembles of INNE \citep{bandaragoda2014efficient} and ADERH \citep{durani2025anomaly}. Other approaches use local neighborhood structure, including $k$-nearest neighbor scoring \citep{bergman2020deep} and local outlier factor (LOF) ~\citep{breunig2000lof}, or fitted feature-space distributions 
\citep{rippel2021modeling}. These methods are often effective, but their performance depends on certain implicit assumptions: isolation methods seek points that are easy to separate based on some heuristic, local methods depend on a chosen neighborhood scale, and distributional methods rely on the adequacy of the fitted density model. On the other hand, deep unsupervised detectors can learn more flexible notions of abnormality \citep{xu2023deep,xu2023fascinating,goodge2022lunar}, but they require training a model for each dataset, which adds computational cost and other deployment-related challenges. 
How can we develop AD methods that retain the efficiency of classical detectors while making fewer assumptions about the geometry of embedding data?

\begin{figure*}
    \centering
\includegraphics[trim={0cm 0 0 0},clip, width=1.\linewidth]{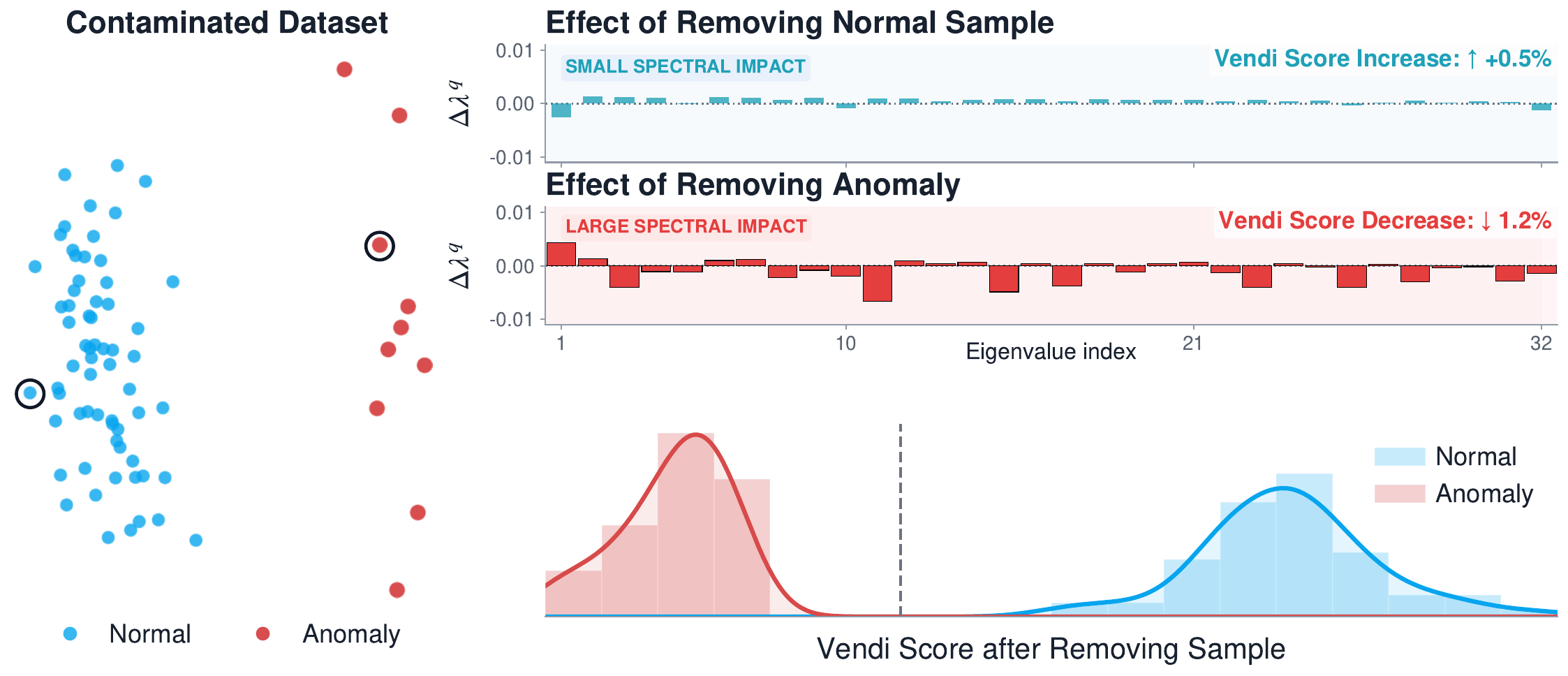}
    \caption{Overview of VAS. Left: A multimodal dataset contaminated with an anomaly. Right: Removing a rare anomalous point induces a larger change in the eigenvalue spectrum than removing a normal sample. VAS detects these shifts by estimating the change in Vendi Score, derived from the Rényi entropy of order $q$ of the spectrum (Eq. \ref{eq:vs-general}).
    }
    \label{fig:Fig1}
\end{figure*}

\textbf{Contribution.} In this paper, we answer this question and make several contributions and findings that we summarize below.
\begin{itemize}
    \item We introduce a new paradigm for AD based on diversity. We propose the Vendi Anomaly Score (VAS), 
    which builds on the Vendi Scores (VS), a family of similarity-based diversity metrics \citep{friedman2022vendi, pasarkar2023cousins}, and defines anomaly through changes in the eigenvalue spectrum of the embedding similarity matrix (Figure~\ref{fig:Fig1}).
    \item We make VAS practical and easy to deploy. We derive an efficient approximation to the exact leave-one-out score, allowing VAS to scale to large embedding datasets. With a simple heuristic, our method also automatically adjusts to different data geometries, eliminating the need for neighborhood selection, fitting clusters, or estimating parametric densities. 
    \item We demonstrate the effectiveness of VAS across diverse embedding datasets and large-scale contamination experiments. On ADBENCH image and text benchmarks \citep{han2022adbench}, spanning $148$ total datasets, VAS outperforms all other baselines. We also evaluate VAS on controlled ImageNet experiments with up to \(100\mathrm{k}\) samples \citep{deng2009imagenet}, showing that VAS is especially effective for detecting contaminating subpopulations at scale.
\end{itemize}

\section{Related Work}
Our work relates to three lines of research, which we describe below.

\textbf{Unsupervised anomaly detection.} Existing unsupervised AD methods typically identify anomalies based on criteria such as separability, local density, or reconstruction error. 

Partitioning-based methods isolate anomalies by recursively subdividing the feature space. Isolation Forest (iForest) \citep{liu2008isolation} and its variants, such as Extended Isolation Forest \citep{hariri2019extended}, identify anomalies as points that can be separated with fewer splits. While effective in low-dimensional or tabular settings, these methods rely on partitions of the feature space, which can struggle to capture complex geometry in high-dimensional embedding spaces. 

A second class of methods estimates local data density. Techniques such as Local Outlier Factor (LOF) \citep{breunig2000lof} and deep nearest neighbors (DN2) \citep{bergman2020deep} identify anomalies as points that lie in low-density regions relative to their neighbors. However, these approaches require constructing neighborhoods of fixed size, making them sensitive to scale and limiting their ability to adapt to heterogeneous or multimodal data distributions.

Ensemble-based methods combine multiple weak signals to improve robustness. For example, Lightweight on-line Detector of Anomalies (LODA) \citep{pevny2016loda} aggregates anomaly scores from random low-dimensional projections, while ADERH leverages ensembles of random hyperspheres to estimate density and support \citep{durani2025anomaly}. ADERH provides the strongest results to date for AD on tabular data, although it requires the strong assumption that normal data can be bounded within hyperspheres, which may not be true for embedding data. 

Recent work has also turned to deep learning to improve AD in complex data. Many of these methods aim to construct representation spaces where anomalies are more easily separated. For example, Deep Isolation Forest \citep{xu2023deep} learns a non-linear transformation of the data that IForest can leverage. 
LUNAR \citep{goodge2022lunar} unifies local outlier methods such as LOF through a graph neural network message-passing framework, introducing learnability into neighborhood-based AD. Similarly, DeepSVDD \citep{ruff2018deep} learns to project normal data within a hypersphere and anomalous data outside of it. RCA \citep{liu2021rca} and RDP \citep{wang2019unsupervised} use collaboration between multiple models to generate a new space where anomalies are more separable. SLAD \citep{xu2023fascinating} and DTE \citep{livernoche2024diffusion} both measure how far a sample deviates from normal data. SLAD learns to predict self-supervised “scale” relationships within each sample and identifies anomalies via inconsistencies in these predictions, while DTE leverages diffusion models to estimate deviation from the normal data manifold.

\textbf{Out-of-distribution detection.} 
Out-of-distribution (OOD) detection is closely related to semi-supervised AD: both settings assume access to in-distribution data and aim to identify samples that deviate from it. In OOD detection, however, the goal is typically to determine whether a test input falls outside the training distribution of a particular predictive model. As a result, many OOD methods are designed around signals produced by that model, including prediction confidence \citep{hendrycks2016baseline}, energy-based scores \citep{liu2020energy,liu2023gen}, feature-space statistics such as Mahalanobis distance \citep{lee2018simple,mueller2025mahalanobis++}, or other properties of the learned representation \citep{fang2025kernel,bergman2020deep, wang2022vim}. Standardized benchmarks such as OpenOOD \citep{yang2022openood} evaluate these methods in model-centric settings where the detector is built around a classifier trained on a specified in-distribution dataset.

In contrast, we study unsupervised AD. Here, the goal is to identify anomalous samples within the dataset being analyzed, using embeddings produced by a model trained independently on that dataset. We do not assume labels, a clean reference set, prediction logits, or model-specific confidence scores. Our method instead operates directly on the similarity structure of the embedding dataset.

\textbf{Measuring rarity with the Vendi Scores.}
\cite{friedman2022vendi} introduced the Vendi Score (VS) as a similarity-based diversity measure. It has since been used to evaluate diversity in a wide variety of domains \citep{stein2023exposing, neshat2025smart, zamzmi2025scorecard, denadel2025evaluating, nielsen2025applications, barcelo2024avoiding, tom2026beyond}. Recent methods have explored how the Vendi Scores can be used to identify rare or novel samples in various settings. In active search, \cite{nguyen2024quality} measure incremental changes to quality-weighted extensions of the VS, showing that changes in the VS can be used to guide the selection of diverse and informative samples in experimental design settings. Similarly, \cite{askari2024improving} leverage incremental changes to the VS to guide image generation toward underrepresented regions of the output space. The VS has also been used for OOD detection: the Vendi Novelty Scores (VNS) detect OOD samples by measuring how much a test sample increases the diversity of a reference dataset by class-conditional novelty estimates derived from a trained classifier \citep{pasarkar2026vendi}. VNS relies on incremental changes to diversity, but it is designed for inference-time OOD detection relative to a fixed training distribution and leverages model-dependent signals. 

\section{Method}
\label{sec:method}
We review the Vendi Scores (VS), and introduce our method for leveraging them in unsupervised AD.

\textbf{The Vendi Scores.} Let $\bX \in \mathbb{R}^{N \times D}$ denote a dataset of $N$ samples. Given a positive semidefinite kernel $k(\cdot,\cdot)$, we construct the similarity matrix $\bK \in \mathbb{R}^{N \times N}$ with entries $K_{ij} = k(x_i, x_j)$. The VS \citep{friedman2022vendi} measures the diversity of $\bX$ via the entropy of the eigenvalues of $\bK$, normalized such that $\text{Tr}(\bK)=1$. 

This definition can be generalized using the \text{R\'enyi} entropy \citep{pasarkar2023cousins}. Let $\lambda_i(\bK)$ denote the $i$th eigenvalue of $\bK$ and $q\geq0$ be the order of the VS. The order-$q$ is defined as 
\begin{equation}
\label{eq:vs-general}
 \text{VS}_q(\bK) = \exp\left(\frac{1}{1-q}\log \sum_{i=1}^{N} \lambda_{i}(\bK)^q\right).
\end{equation}
This simplifies to the exponential Shannon Entropy when $q=1$. The parameter $q$ controls sensitivity to different parts of the spectrum: smaller values emphasize the tail, while larger values emphasize the leading eigenmodes \citep{pasarkar2023cousins}.

Computing $\mathrm{VS}_q$ naively requires the eigendecomposition of the $N \times N$ matrix $\bK$, which requires $O(N^3)$ complexity. However, as \cite{friedman2022vendi} note, when working with $\ell_2$-normalized features and using a cosine kernel, the VS can be computed more efficiently. In this setting, the similarity matrix is $\bK = XX^T/N$ and has the same non-zero eigenvalues of the smaller $D \times D$ matrix $X^TX/N$. We therefore can compute the spectrum using $X^TX/N$, reducing the computational cost to $O(D^2N + D^3)$, which is considerably more efficient in the common ML setting where the number of samples greatly exceeds the feature dimension ($N \gg D$).  

\textbf{Fast estimation of diversity influence for anomaly detection.}
We consider a dataset of $N$ samples represented by $D$-dimensional embeddings $\bX \in \mathbb{R}^{N \times D}$, obtained from a pretrained model. The goal of unsupervised AD is to identify which samples are anomalous relative to the rest of the dataset. We aim to learn a scoring function $s : \{1, \dots, N\} \rightarrow \mathbb{R}$ that assigns a real-valued anomaly score to each sample based on its embedding. Higher scores correspond to more anomalous points. Only the induced ordering matters: any transformation of $s(i)$ that preserves the ordering yields an equivalent solution.

We define anomaly scores based on the change in dataset diversity when a sample is removed. Assume $\ell_2$ normalized features and let
$\rho = \frac{1}{N} \bX^\top \bX \in \mathbb{R}^{D \times D}$
denote the feature covariance matrix corresponding to the cosine kernel. For a sample $x_i \in \mathbb{R}^D$, let $\bX_{-i}$ denote the dataset with $x_i$ removed, and define
$\rho_{-i} = \frac{1}{N-1} \bX_{-i}^\top \bX_{-i}
= \frac{\bX^\top \bX - x_i x_i^\top}{N-1}$.
We then define the anomaly score as, 
\begin{align*}
\Delta(x_i)
=
\log \mathrm{VS}_q(\rho) - \log \mathrm{VS}_q(\rho_{-i}).
\end{align*}

Intuitively, samples that contribute uniquely to the structure of the dataset will induce a larger decrease in diversity when removed, yielding larger values of $\Delta(x_i)$, as shown in Figure \ref{fig:Fig1}. 

However, computing $\Delta(x_i)$ naively for all $N$ samples requires $O(N D^3)$ time, which is prohibitive for high-dimensional datasets. Moreover, this formulation requires recomputing the spectrum of $\rho_{-i}$ after removing each sample, preventing efficient vectorization and limiting the ability to leverage batched GPU computation. We can accelerate this computation by noting that each removal corresponds to a rank-one perturbation of $\rho$ of size $O(1/N)$. Proposition~\ref{prop:first-order-vas} shows that the resulting change in Vendi Score can be approximated by a single quadratic form involving $x_i$. 

\begin{proposition}[First-order approximation in log-space]
\label{prop:first-order-vas}
Assume that \(\log \mathrm{VS}_q\) is twice continuously differentiable in a neighborhood of \(\rho\). Let
\[
G_q =
\begin{cases}
\displaystyle
\frac{q}{(1-q)\operatorname{Tr}(\rho^q)} \rho^{q-1}, & q \neq 1, \\[1.2em]
\displaystyle
-(\log \rho + I), & q = 1.
\end{cases}
\]
Then, for each \(i\),
\[
\log \mathrm{VS}_q(\rho) - \log \mathrm{VS}_q(\rho_{-i})
=
\frac{x_i^\top G_q x_i - \operatorname{Tr}(G_q \rho)}{N-1}
+
O\!\left(\frac{1}{N^2}\right).
\]
\end{proposition}

A proof is provided in Appendix Section \ref{sec:proof}. The differentiability assumption excludes spectral boundary cases in which zero eigenvalues make terms such as $\log \rho$ or $\rho^{q-1}$ singular. In implementation, all spectral quantities are computed on the positive eigenspace of $\rho$, equivalently restricting the calculation to the span of the data. 

Because $\operatorname{Tr}(G_q \rho)$ in Proposition \ref{prop:first-order-vas} is constant across samples in a dataset, and we only care about ranking samples from our scoring function, we only need to compute the quantity $x_i^\top G_q x_i$ for each sample $x_i$. This yields a simple batched computation of the VAS: 
\begin{align}
    \label{eqn:vas}
    \text{VAS}(x_i):= \operatorname{diag}(\bX G_q \bX^\top)_i \qquad \forall i=1\dots N.
\end{align}
We empirically demonstrate the accuracy of the approximation used in VAS in Section \ref{subsec:empirical_acc}. 

VAS can also be interpreted as a spectral quadratic anomaly score, connecting it to classical methods such as Mahalanobis distance and PCA reconstruction error. While these methods use predefined rules for weighting different directions of variation, VAS derives its weighting from each sample's influence on dataset diversity, with the relative importance of different directions controlled by the Vendi Score order \(q\). In Appendix Section \ref{sec:spectral_comparison}, we provide additional description of how VAS relates to other spectral scores. 

\textbf{Automatically adapting $q$.} VAS introduces a single hyperparameter: the order $q$ of the Vendi Score. Previous methods leveraging the Vendi Scores often fix a single value of $q$ across experiments \citep{yang2025measuring, pasarkar2026vendi} or track the VS across a range of q values to draw diversity profiles \citep{dieng2025unified, hu2026evaluating}. However, in unsupervised AD, the optimal choice of $q$ can vary significantly depending on the geometry of the embedding space, and we are not afforded a validation set to tune hyperparameters. Therefore, we design a simple, data-dependent heuristic that adapts $q$ automatically across datasets.

The parameter $q$ controls how strongly the Vendi Score emphasizes the tail of the eigenvalue spectrum. The importance of this tail varies across datasets. In embedding spaces where the data is highly concentrated, such as BERT embeddings of text, most variance is captured by the leading eigenmode, resulting in a large principal eigenvalue $\lambda_1$ of the trace-normalized similarity matrix. In this regime, AD requires amplifying the contribution of the spectral tail, which is achieved by using smaller values of $q$. In contrast, in more isotropic embedding spaces, such as image representations spanning diverse visual features, the variance is more evenly distributed, leading to a smaller $\lambda_1$. In these settings, less emphasis on the tail is needed, and larger values of $q$ are appropriate.

To capture this behavior without manual tuning, we define $q$ as a function of the principal eigenvalue:
\begin{align*}
    q = \max\big(2(1 - \lambda_1),\, 0.1\big).
\end{align*}
The lower bound of $0.1$ is introduced for numerical stability when evaluating $\rho^{q-1}$ for small $q$. This heuristic assigns smaller values of $q$ to concentrated datasets and larger values to more dispersed ones, adapting automatically to the spectral structure of the data.

\textbf{Interpreting VAS.}
Proposition~\ref{prop:first-order-vas} shows that VAS ranks samples by the quadratic form \(x^\top G_q x\). Therefore, to understand what kinds of subpopulations VAS identifies as anomalous, it suffices to study the expected
score \(\mathbb E_{x\sim P_c}[x^\top G_qx]\) for samples drawn from a component \(P_c\). Proposition~\ref{prop:vas_component_spectral_density} analyzes this quantity under a stylized orthogonal mixture model.


\begin{proposition}[Expected VAS under orthogonal mixtures]
\label{prop:vas_component_spectral_density}
Let \(P=\sum_{c=1}^K \pi_c P_c\) be a mixture distribution on
\(\mathbb S^{d-1}\), and define $\rho=\mathbb E_{x\sim P}[xx^\top]$. 
Assume each component \(P_c\) is supported on a subspace
\(U_c\subseteq \mathbb R^d\), and that the subspaces \(U_1,\ldots,U_K\) are
mutually orthogonal. Let $\rho_c=\mathbb E_{x\sim P_c}[xx^\top]$. 
Using $G_q$ as defined in Proposition \ref{prop:first-order-vas}, for any \(x\in U_c\) and \(q\neq1\),
\[
x^\top G_qx
=
\frac{q}{(1-q)\operatorname{Tr}(\rho^q)}
\pi_c^{q-1}
x^\top \rho_c^{q-1}x.
\]
Consequently, the expected VAS score of a sample from component \(c\) is
\[
\mathbb E_{x\sim P_c}[x^\top G_qx]
=
\frac{q}{(1-q)\operatorname{Tr}(\rho^q)}
\left(
\frac{\pi_c}{\mathrm{VS}_q(\rho_c)}
\right)^{q-1}.
\]
\end{proposition}
The proof is provided in Appendix \ref{sec:vas_component_spectral_density}. Proposition~\ref{prop:vas_component_spectral_density} shows that the expected VAS score for samples from component \(c\) depends on the ratio \(\pi_c/\mathrm{VS}_q(\rho_c)\). Here, \(\pi_c\) is the component mass, while \(\mathrm{VS}_q(\rho_c)\) is the component's effective spectral dimension. Thus, VAS is sensitive to how much probability mass a component has relative to its internal diversity. Components that are small but spectrally diffuse can receive high anomaly scores, even if their samples are not isolated from one another. This component-level sensitivity emerges without explicitly identifying components or forming local neighborhoods; it arises implicitly from the spectrum of the full dataset.
\newcommand{\smallscript}{\fontsize{6.0}{7.5}\selectfont}

\begin{table*}[t]
\scriptsize
\setlength{\tabcolsep}{2.2pt}
\begin{tabular}{llcccccccccccc}
\toprule
Dataset & Encoder & VAS & ADERH & DIF & IForest & LUNAR & DN2 & LODA & LOF & KPCA & SLAD & GMD & DTE \\
\midrule

CIFAR10 & ResNet18
& \second{0.745} & 0.633 & 0.684 & 0.627 & 0.657 & \third{0.740} & 0.635 & 0.696 & \best{0.751} & 0.735 & 0.732 & 0.673 \\

CIFAR10 & ViT
& \second{0.941} & \third{0.924} & 0.860 & 0.876 & 0.837 & 0.855 & 0.856 & 0.519 & 0.905 & \best{0.942} & 0.898 & 0.590 \\

FashionMNIST & ResNet18
& \third{0.911} & 0.865 & 0.868 & 0.831 & 0.852 & 0.858 & 0.829 & 0.605 & \best{0.915} & 0.904 & \second{0.912} & 0.848 \\

FashionMNIST & ViT
& \best{0.930} & 0.910 & 0.856 & 0.890 & 0.815 & 0.860 & 0.862 & 0.519 & 0.917 & \third{0.919} & \second{0.925} & 0.657 \\

MNIST-C & ResNet18
& \second{0.793} & 0.763 & 0.744 & 0.734 & 0.780 & 0.763 & 0.724 & 0.628 & 0.765 & \third{0.782} & \best{0.793} & 0.759 \\

MNIST-C & ViT
& \best{0.739} & 0.723 & 0.704 & 0.720 & 0.695 & 0.704 & 0.689 & 0.555 & 0.719 & \second{0.732} & \third{0.723} & 0.568 \\

MVTec-AD & ResNet18
& \best{0.799} & 0.755 & 0.750 & 0.745 & 0.768 & \second{0.797} & 0.704 & 0.763 & 0.741 & 0.788 & \third{0.790} & 0.705 \\

MVTec-AD & ViT
& \best{0.896} & 0.809 & 0.795 & 0.808 & 0.858 & \third{0.875} & 0.781 & 0.842 & 0.847 & 0.851 & \second{0.891} & 0.697 \\

SVHN & ResNet18
& \third{0.657} & 0.579 & 0.615 & 0.576 & 0.609 & \best{0.662} & 0.589 & 0.638 & 0.657 & 0.655 & \second{0.659} & 0.592 \\

SVHN & ViT
& \third{0.579} & 0.519 & 0.524 & 0.530 & 0.564 & \best{0.584} & 0.524 & 0.577 & 0.565 & 0.553 & \second{0.580} & 0.529 \\

20News & BERT
& \best{0.626} & 0.543 & 0.557 & 0.555 & 0.571 & 0.565 & 0.554 & \second{0.607} & 0.554 & \third{0.606} & 0.603 & 0.560 \\

20News & RoBERTa
& \best{0.749} & 0.636 & 0.624 & 0.643 & 0.698 & \second{0.732} & 0.607 & \third{0.723} & 0.720 & 0.703 & 0.716 & 0.503 \\

Agnews & BERT
& \second{0.676} & 0.583 & 0.582 & 0.583 & 0.650 & 0.610 & 0.558 & \best{0.725} & 0.604 & \third{0.656} & 0.650 & 0.587 \\

Agnews & RoBERTa
& \best{0.881} & 0.718 & 0.661 & 0.677 & 0.790 & \second{0.850} & 0.638 & 0.719 & \third{0.846} & 0.797 & 0.813 & 0.501 \\

Amazon & BERT
& \second{0.608} & 0.577 & 0.565 & 0.558 & \best{0.608} & 0.595 & 0.531 & 0.578 & 0.585 & \third{0.606} & 0.604 & 0.522 \\

Amazon & RoBERTa
& 0.544 & 0.485 & 0.518 & 0.509 & 0.536 & \best{0.551} & 0.520 & 0.548 & \second{0.551} & 0.532 & 0.530 & \third{0.548} \\

Imdb & BERT
& 0.502 & 0.486 & 0.490 & 0.488 & 0.493 & \second{0.517} & 0.447 & 0.505 & \best{0.517} & 0.513 & \third{0.515} & 0.483 \\

Imdb & RoBERTa
& \second{0.530} & 0.439 & 0.508 & 0.485 & 0.518 & 0.529 & 0.506 & \best{0.532} & \third{0.530} & 0.510 & 0.504 & 0.500 \\

Yelp & BERT
& \best{0.679} & 0.623 & 0.617 & 0.603 & 0.664 & \third{0.672} & 0.573 & 0.667 & 0.647 & 0.659 & \second{0.676} & 0.467 \\

Yelp & RoBERTa
& 0.566 & 0.511 & 0.537 & 0.531 & \third{0.578} & \best{0.594} & 0.548 & \second{0.581} & 0.577 & 0.568 & 0.575 & 0.532 \\

\midrule
Average Rank
& & \best{2.35} & 8.70 & 8.55 & 9.20 & 6.10 & \third{4.05} & 10.15 & 6.25 & 4.70 & 4.35 & \second{3.80} & 9.80 \\
p-Value
& & - & \smallscript{1.34e-20} & \smallscript{1.15e-24} & \smallscript{6.27e-24} & \smallscript{3.78e-22} & \smallscript{3.14e-10} & \smallscript{2.08e-25} & \smallscript{5.23e-15} & \smallscript{4.11e-15} & \smallscript{4.49e-16} & \smallscript{1.34e-08} & \smallscript{2.07e-24} \\
\bottomrule
\end{tabular}
\caption{AUROC across dataset families and encoders. For multi-split datasets, we report average AUROC. \best{Orange} marks best, \second{blue} marks 2nd. Bottom-row p-values are paired Wilcoxon tests: VAS vs. each baseline over all 148 dataset-split/encoder instances.}
\label{tab:AUROC_Main}
\end{table*}

\textbf{Time and space complexity.} 
The computational cost of VAS is dominated by forming
$\rho = \bX^\top \bX/N$, eigen-decomposing the resulting $D \times D$ matrix, and computing the VAS scores from Equation \ref{eqn:vas} for all
samples. Forming $\rho$ costs $O(ND^2)$ time, the eigendecomposition costs
$O(D^3)$ time, and the batched score computation costs $O(ND^2)$ time. Thus, the total runtime is $O(ND^2 + D^3)$, with memory complexity
$O(ND + D^2)$. In contrast, naively recomputing the Vendi Score after each sample removal would require $N$ separate eigendecompositions, yielding $O(ND^3)$ time complexity. VAS therefore avoids per-sample spectral computations and
reduces scoring to a single global eigendecomposition followed by batched matrix operations.

\section{Experiments}
In this section, we compare VAS against a diverse set of AD algorithms. We benchmark all algorithms on various image and text embedding datasets across multiple architectures. 

\textbf{Datasets and models.} We evaluate VAS on the embedding-based image and text datasets from ADBENCH \citep{han2022adbench}. These datasets span numerous image and text datasets and architectures. For all image datasets, we test methods against embeddings from both ResNet18 \citep{he2016deep} and ViT \citep{dosovitskiy2020image} architectures that had been pre-trained on ImageNet \citep{deng2009imagenet}. For text data, we test performance on embeddings from BERT \citep{devlin2019bert} and RoBERTa \citep{liu2019roberta}. The benchmark covers a wide range of AD scenarios. These include rare class detection (e.g., CIFAR-10 \citep{krizhevsky2009learning}, SVHN  \citep{netzer2011reading}), where one class is treated as normal and samples from the remaining classes are downsampled to form anomalies; distributional corruption (e.g., MNIST vs. MNIST-C \citep{mu2019mnist}), where anomalies arise from shifts in data generation; and semantic or structural anomalies (e.g., MVTec-AD \citep{bergmann2019mvtec}). In text datasets (e.g., Amazon \citep{he2016ups}, Yelp \citep{zhang2015character}), anomalies are constructed via shifts in semantic content, such as contrasting sentiment. Additional specifications for each dataset are provided in Appendix Section \ref{sec:datasest}.

We additionally perform AD experiments on ImageNet-1K \citep{deng2009imagenet}, using embeddings extracted from the DinoV3 ViT-B/16 model \citep{simeoni2025dinov3}.

\textbf{Baselines.}
We compare against $11$ unsupervised AD algorithms:
ADERH \citep{durani2025anomaly}, DIF \citep{xu2023deep}, IForest \citep{liu2008isolation}, DN2 \citep{bergman2020deep}, LODA \citep{pevny2016loda}, LOF \citep{breunig2000lof},
KPCA \citep{fang2025kernel}, SLAD \citep{xu2023fascinating}, DTE \citep{livernoche2024diffusion}, and LUNAR \citep{goodge2022lunar}. We also include a custom Gaussian-mixture Mahalanobis distance baseline
(GMD) to represent fitted-distribution methods. Inspired by Gaussian feature
modeling for AD \citep{rippel2021modeling}, GMD fits a
Bayesian Information Criterion-selected Gaussian mixture model to the unlabeled embeddings and scores each sample by its Mahalanobis distance to its most likely component. Together, these baselines span feature-space partitions (IForest), local neighborhoods (LOF, DN2), randomized ensemble methods based on projections or hyperspheres (LODA, ADERH), deep anomaly detectors (SLAD, DIF, DTE, LUNAR), kernel reconstruction methods (KPCA), and fitted feature-space distributions (GMD). Additional descriptions of each algorithm and their relationship to VAS are provided in Appendix Section~\ref{sec:baselines}.

\textbf{Hyperparameters and preprocessing.} All main-text experiments use the default hyperparameter recommendations from each method's reference implementation. VAS uses the adaptive choice of $q$ described in Section~\ref{sec:method}, with no validation-label tuning. To assess sensitivity of baselines to the default hyperparameter protocol, we additionally report a grid-search comparison in Appendix Section~\ref{sec:grid-search}.

For preprocessing, we use a method-specific protocol designed to match the assumptions of each baseline rather than imposing a single uniform pipeline.
This avoids disadvantaging methods whose scoring rules depend on a particular feature scale. Specifically, ADERH uses \texttt{MinMaxScaler} and LUNAR uses \texttt{StandardScaler} \citep{pedregosa2011scikit}. SLAD, DIF, IForest, DTE, LOF, and LODA are applied to raw features, consistent with their standard usage. DN2 and GMD operate on $\ell_2$-normalized features, following recent evidence that normalized embeddings improve distance- and Mahalanobis-based scoring in deep feature spaces \citep{sun2022out,mueller2025mahalanobis++}. Finally, KPCA and VAS use a cosine kernel, so they operate on $\ell_2$-normalized features.

\textbf{Evaluation protocol.}
All methods are fit directly on the contaminated unlabeled dataset and then used to score the same samples, matching the standard unsupervised AD setting. We evaluate performance using Area Under the ROC Curve (AUROC) and Average Precision (AP). Higher values indicate better performance for both metrics.

\begin{figure}
    \centering
    \includegraphics[width=0.95\linewidth]{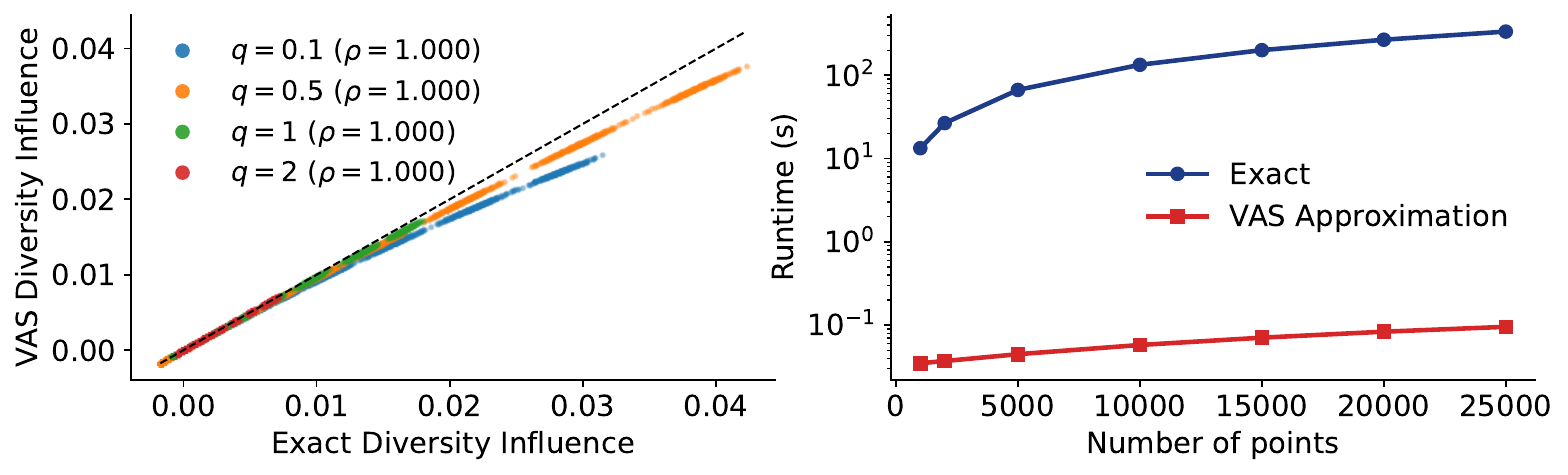}
    \caption{VAS provides accurate and efficient diversity influence measurements. Left: VAS diversity influence compared against the exact leave-one-out Vendi Score change for multiple orders of $q$. Spearman rank correlation $\rho$ is shown for each choice of $q$. Right: Varying dataset sizes shows that VAS provides orders of magnitude speedup over the exact approach.}
    \label{fig:approx_acc}
\end{figure}

\subsection{Empirical Accuracy of VAS}
\label{subsec:empirical_acc}
Prior to benchmarking VAS against other baselines, we verify the accuracy of its approximation to the exact leave-one-out VS change. In Figure \ref{fig:approx_acc}, we test the method on a $25,000$-point, $100$-component mixture of gaussians dataset in $512$ dimensions. For all tested VS orders $q$, we achieve a Spearman rank correlation of $\rho=1.00$, showing that the approximation preserves the sample ranking. We also show the preferred efficiency of VAS: Varying the dataset size from $1000$ to $25000$ points shows that VAS is orders of magnitude faster than the exact approach.

We additionally evaluate the approximation on all 148 ADBENCH benchmark instances by comparing VAS against the exact leave-one-out Vendi Score change. The approximation closely preserves the anomaly rankings, with an average Spearman $\rho$ over 0.98 on all tested benchmarks. The change in AUROC is at most $1.16\%$. Detailed results are provided in Table \ref{tab:approx_adbench}.

\subsection{ADBENCH Anomaly Detection Performance Results}
Tables \ref{tab:AUROC_Main} and \ref{tab:AP_main} show the performance of VAS relative to other baselines on the ADBENCH embedding benchmarks based on AUROC and AP, respectively. We report family-level averages across datasets and across the two embedding models available for each benchmark family. Statistical tests are also run across all $148$ benchmark instances, where each instance corresponds to a dataset split paired with an embedding model. VAS achieves significantly stronger performance than the competing methods under a Wilcoxon signed-rank test for both AUROC and AP.

Figure \ref{fig:q_histogram} shows the distribution of the adaptively selected Vendi orders $q$. The order $q$ controls which parts of the spectrum drive the diversity-influence score: larger values emphasize dominant eigen-directions, while smaller values place relatively more weight on lower-variance directions. We observe that NLP embeddings tend to have larger principal eigenvalues than CV embeddings, suggesting stronger concentration around shared semantic context. Consequently, the adaptive rule selects smaller values of $q$ for NLP datasets, which makes VAS more sensitive to variation in the spectral tail.

To ensure that our comparisons are robust against hyperparameters, we perform a grid search over the hyperparameters of all baseline methods. For each method, we pick the hyperparameters that give the strongest performance relative to VAS. Even in this setting of letting methods overfit to the evaluation sets, VAS still provides the strongest results (Tables \ref{tab:auroc_sweep} and \ref{tab:ap_sweep}). Additional details of this grid search experiment are in Appendix Section \ref{sec:grid-search}. 

We additionally evaluate the sensitivity of VAS to alternative rules for selecting $q$, including different scaling factors, multiple leading eigenvalues, and the effective spectral rank. Our heuristic performs comparably to several alternatives and significantly outperforms others across 148 ADBENCH tasks, suggesting that its effectiveness is not overly sensitive to the choice of heuristic (Appendix~\ref{sec:choice_of_q}).

\begin{figure*}
    \centering
\includegraphics[width=1.\linewidth]{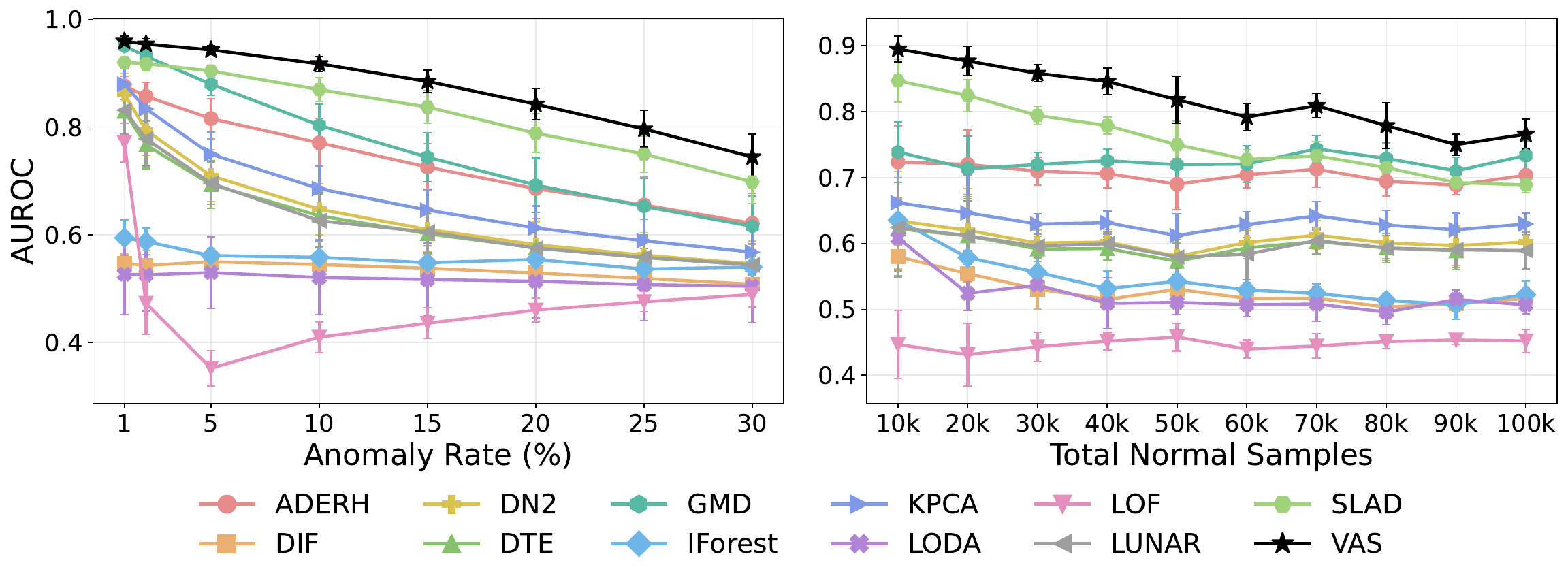}
    \caption{VAS achieves accurate ImageNet AD. We construct contaminated datasets by sampling normal and anomalous images from disjoint ImageNet classes. Left: AUROC as the anomaly rate varies with $25{,}000$ normal images from $25$ classes. Right: AUROC as dataset size increases at a fixed $20\%$ anomaly rate, adding normal samples in increments of $10,000$. VAS performs best across both sweeps, showing robustness to increasing contamination and scale. Results are averaged over $5$ replications, with standard deviation bars included.
    }
    \label{fig:imagenet}
\end{figure*}

\subsection{ImageNet Anomaly Detection Results}
We next evaluate VAS in a controlled large-scale image setting using ImageNet embeddings from DINOv3 ViT-B/16 \citep{simeoni2025dinov3}. We construct contaminated datasets by sampling normal images from one set of ImageNet classes and anomalous images from a disjoint set of classes. We study two settings. First, we fix the normal set to $25{,}000$ images from $25$ classes and vary the number of anomalous images sampled from $25$ disjoint classes,
thereby sweeping the anomaly rate. Second, we fix the anomaly rate at $20\%$ and scale the dataset by increasing the number of normal classes in increments of $10$, while sampling anomalies from an equal number of disjoint classes. This increases both dataset size and semantic diversity as we scale to $100{,}000$ normal images. Across both settings, VAS uses the same adaptive choice of $q$ as in the ADBENCH experiments and achieves the strongest performance among the evaluated methods (Fig.~\ref{fig:imagenet}). SLAD provides the most competitive performance, although as shown in the following section, it is orders of magnitude slower than VAS.

These results match the behavior predicted by
Proposition~\ref{prop:vas_component_spectral_density}. The anomalous samples in this experiment are not isolated outliers: they are sampled from a select number of ImageNet classes and can therefore form coherent local neighborhoods in feature space. This structure can weaken
local density methods, which may treat anomalous samples as typical once they have nearby anomalous neighbors. VAS instead scores each sample by its diversity influence. Under the mixture structure described in Proposition~\ref{prop:vas_component_spectral_density}, this score depends on both component mass and internal diversity, enabling VAS to detect rare but coherent anomalous subpopulations.

To further investigate VAS across different anomaly types, we consider three additional constructions using ADBENCH embeddings: locally class-confusable observations, global directional perturbations, and disrupted feature dependencies. VAS performs well across constructions compared to the other best-performing baselines. We provide details of these experiments in Appendix~\ref{sec:anomaly_types}.

\subsection{Latency Evaluation of VAS}
\begin{figure*}
    \centering
\includegraphics[width=1.\linewidth]{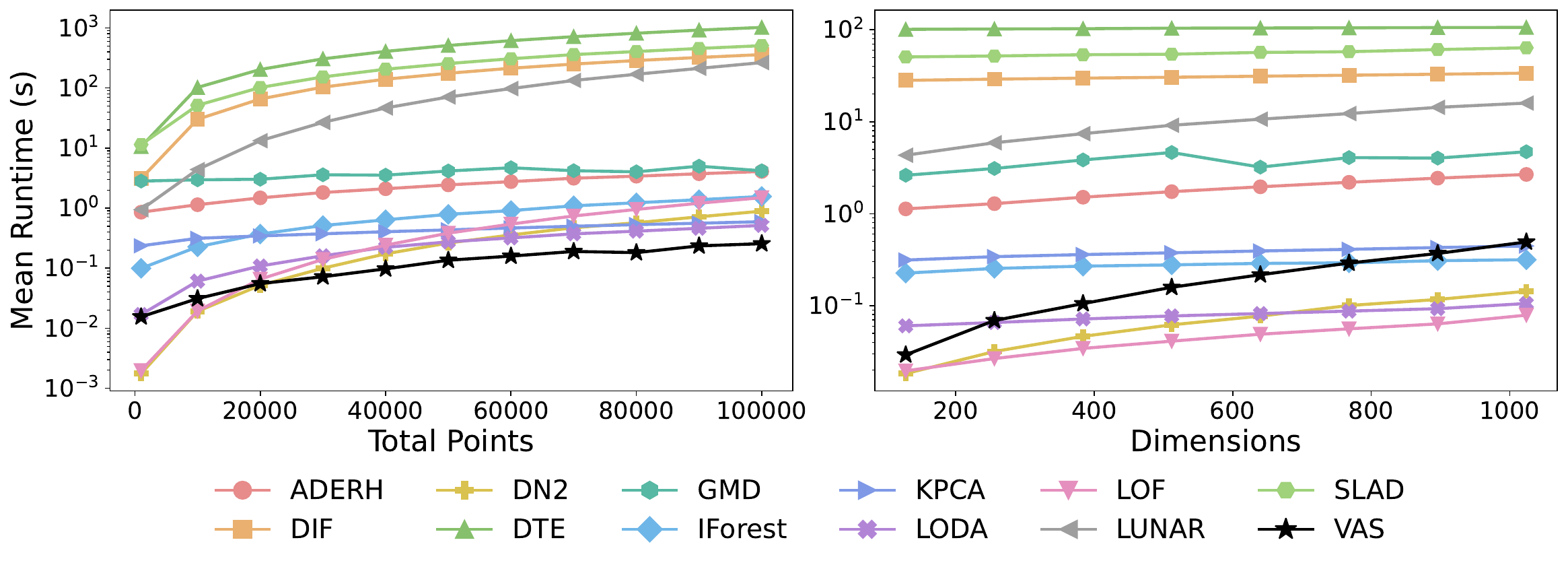}
    \caption{VAS supports efficient AD. Left: Latency of methods when increasing dataset size at a fixed dimensionality of $128$. Right: Latency of methods with increasing dimensionality at a fixed dataset size of $5000$. Results are averaged over $n=3$ repetitions.}
    \label{fig:latency}
\end{figure*}

We compare the latency of VAS and all tested baselines on simulated Gaussian datasets while varying either the number of samples or the embedding dimension (Figure \ref{fig:latency}). We report the time required to fit each method and score all samples. All experiments were performed on a single NVIDIA A6000 GPU. For ADERH, LODA, and LOF, we use custom GPU-friendly implementations to provide fairer latency comparisons, and verify that their outputs match the original implementations.

Figure \ref{fig:latency} shows that neural baselines are substantially slower across the tested regimes. As the number of samples increases, VAS exhibits linear scaling and becomes one of the fastest methods on larger datasets. This follows from its batched quadratic-form computation after the matrix $G_q$ has been formed. VAS has a stronger dependence on embedding dimension than alternatives such as IForest or LODA. However, in embedding-based AD, the dimension is fixed by the pretrained model. Thus, the more relevant large-scale regime is increasing $N$ at fixed $D$, where VAS scales linearly.

\section{Conclusion}
This work introduces \textbf{VAS}, an unsupervised AD method for embedding data that scores each sample by its influence on dataset diversity. By deriving a first-order approximation to leave-one-out Vendi Scores, VAS avoids repeated eigendecompositions, requires no validation-based hyperparameter tuning, and adapts automatically to the geometry of the dataset. Across image and text AD benchmarks, VAS consistently improves over classical, kernel-based, fitted-distribution, and deep AD baselines. Our large-scale ImageNet experiments further show that VAS is especially effective for detecting coherent contamination, where anomalous samples form rare but structured subpopulations. Together, these results establish VAS as a scalable and effective approach to unsupervised AD. 

\textbf{Limitations.}
VAS depends on the quality of the embedding space and the chosen similarity function. If anomalies are indistinguishable from normal samples under the embedding or cosine similarity, VAS cannot reliably detect them. While VAS provides a principled anomaly ranking, the scalar score does not by itself explain the source of anomalousness. In addition, our adaptive rule for selecting \(q\) is heuristic and may not be optimal for all data geometries.

\textbf{Future work.}
Future work includes extending VAS to learned or task-specific kernels and developing principled rules for adapting the VS order to dataset geometry. More broadly, VAS may be useful for other data curation tasks, including coreset selection and duplicate detection.

\textbf{Broader impacts.}
VAS can support dataset auditing by helping identify corrupted or mislabeled samples in large embedding datasets. However, anomaly scores depend on the embedding model and similarity function, and may reflect their biases. In sensitive applications, VAS should therefore be used as a tool for prioritizing human review rather than as an automatic rule for removing data.

\begin{ack}
Adji Bousso Dieng acknowledges support from the National Science Foundation CAREER grant CAIF \#2541965. Amey Pasarkar is supported by an NSF-GRFP fellowship. The authors declare no competing interests.
\end{ack}

\paragraph{AI use statement.} In this work, we used generative AI tools for implementing methods in code. We have not used generative AI tools for generating data sets, developing theoretical models or conceptual frameworks, formulating mathematical claims, assisting in the writing of proofs, proposing hypotheses, cleaning and reformatting datasets, supporting qualitative and thematic data analysis, interpreting results or designing or providing feedback on research methodology. We have reviewed all AI-assisted work: All code went through testing to ensure we provided faithful reproductions of established baselines and that our code matched the experimental descriptions in the paper. AI tools were used as a small portion of the paper editing process and all changes were reviewed by the Authors. We take responsibility for the final content of this work, including text, claims, or artifacts produced with the aid of generative AI.

\bibliographystyle{plainnat}
\bibliography{main}

@inproceedings{
durani2025anomaly,
title={Anomaly Detection by an Ensemble of  Random Pairs of Hyperspheres},
author={Walid Durani and Collin Leiber and Khalid Durani and Claudia Plant and Christian B{\"o}hm},
booktitle={The Thirty-ninth Annual Conference on Neural Information Processing Systems},
year={2025},
url={https://openreview.net/forum?id=vM4PIjsJDG}
}

@article{campolongo2025building,
  title={Building machine learning challenges for anomaly detection in science},
  author={Campolongo, Elizabeth G and Chou, Yuan-Tang and Govorkova, Ekaterina and Bhimji, Wahid and Chao, Wei-Lun and Harris, Chris and Hsu, Shih-Chieh and Lapp, Hilmar and Neubauer, Mark S and Namayanja, Josephine and others},
  journal={arXiv preprint arXiv:2503.02112},
  year={2025}
}

@article{chandola2009anomaly,
  title={Anomaly detection: A survey},
  author={Chandola, Varun and Banerjee, Arindam and Kumar, Vipin},
  journal={ACM computing surveys (CSUR)},
  volume={41},
  number={3},
  pages={1--58},
  year={2009},
  publisher={ACM New York, NY, USA}
}

@article{xu2023deep,
  title={Deep isolation forest for anomaly detection},
  author={Xu, Hongzuo and Pang, Guansong and Wang, Yijie and Wang, Yongjun},
  journal={IEEE transactions on knowledge and data engineering},
  volume={35},
  number={12},
  pages={12591--12604},
  year={2023},
  publisher={IEEE}
}

@inproceedings{liu2008isolation,
  title={Isolation forest},
  author={Liu, Fei Tony and Ting, Kai Ming and Zhou, Zhi-Hua},
  booktitle={2008 eighth ieee international conference on data mining},
  pages={413--422},
  year={2008},
  organization={IEEE}
}

@inproceedings{xu2023fascinating,
  title={Fascinating supervisory signals and where to find them: Deep anomaly detection with scale learning},
  author={Xu, Hongzuo and Wang, Yijie and Wei, Juhui and Jian, Songlei and Li, Yizhou and Liu, Ning},
  booktitle={International Conference on Machine Learning},
  pages={38655--38673},
  year={2023},
  organization={PMLR}
}

@inproceedings{breunig2000lof,
  title={LOF: identifying density-based local outliers},
  author={Breunig, Markus M and Kriegel, Hans-Peter and Ng, Raymond T and Sander, J{\"o}rg},
  booktitle={Proceedings of the 2000 ACM SIGMOD international conference on Management of data},
  pages={93--104},
  year={2000}
}

@article{bergman2020deep,
  title={Deep nearest neighbor anomaly detection},
  author={Bergman, Liron and Cohen, Niv and Hoshen, Yedid},
  journal={arXiv preprint arXiv:2002.10445},
  year={2020}
}

@article{fang2025kernel,
  title={Kernel PCA for Out-of-Distribution Detection: Non-Linear Kernel Selections and Approximations},
  author={Fang, Kun and Tao, Qinghua and He, Mingzhen and Lv, Kexin and Yang, Runze and Hu, Haibo and Huang, Xiaolin and Yang, Jie and Cao, Longbin},
  journal={arXiv preprint arXiv:2505.15284},
  year={2025}
}

@article{mueller2025mahalanobis++,
  title={Mahalanobis++: Improving OOD detection via feature normalization},
  author={Mueller, Maximilian and Hein, Matthias},
  journal={arXiv preprint arXiv:2505.18032},
  year={2025}
}

@article{dai2024unsupervised,
  title={Unsupervised anomaly detection for tabular data using noise evaluation},
  author={Dai, Wei and Hwang, Kai and Fan, Jicong},
  journal={arXiv preprint arXiv:2412.11461},
  year={2024}
}

@article{pevny2016loda,
  title={Loda: Lightweight on-line detector of anomalies},
  author={Pevn{\`y}, Tom{\'a}{\v{s}}},
  journal={Machine Learning},
  volume={102},
  number={2},
  pages={275--304},
  year={2016},
  publisher={Springer}
}

@article{hariri2019extended,
  title={Extended isolation forest},
  author={Hariri, Sahand and Kind, Matias Carrasco and Brunner, Robert J},
  journal={IEEE transactions on knowledge and data engineering},
  volume={33},
  number={4},
  pages={1479--1489},
  year={2019},
  publisher={IEEE}
}

@inproceedings{ethayarajh2019contextual,
  title={How contextual are contextualized word representations? Comparing the geometry of BERT, ELMo, and GPT-2 embeddings},
  author={Ethayarajh, Kawin},
  booktitle={Proceedings of the 2019 conference on empirical methods in natural language processing and the 9th international joint conference on natural language processing (EMNLP-IJCNLP)},
  pages={55--65},
  year={2019}
}

@article{friedman2022vendi,
  title={The vendi score: A diversity evaluation metric for machine learning},
  author={Friedman, Dan and Dieng, Adji Bousso},
  journal={arXiv preprint arXiv:2210.02410},
  year={2022}
}

@article{pasarkar2023cousins,
  title={Cousins of the vendi score: A family of similarity-based diversity metrics for science and machine learning},
  author={Pasarkar, Amey P and Dieng, Adji Bousso},
  journal={arXiv preprint arXiv:2310.12952},
  year={2023}
}

@article{han2022adbench,
  title={Adbench: Anomaly detection benchmark},
  author={Han, Songqiao and Hu, Xiyang and Huang, Hailiang and Jiang, Minqi and Zhao, Yue},
  journal={Advances in neural information processing systems},
  volume={35},
  pages={32142--32159},
  year={2022}
}

@article{yang2022openood,
  title={Openood: Benchmarking generalized out-of-distribution detection},
  author={Yang, Jingkang and Wang, Pengyun and Zou, Dejian and Zhou, Zitang and Ding, Kunyuan and Peng, Wenxuan and Wang, Haoqi and Chen, Guangyao and Li, Bo and Sun, Yiyou and others},
  journal={Advances in Neural Information Processing Systems},
  volume={35},
  pages={32598--32611},
  year={2022}
}

@article{hendrycks2016baseline,
  title={A baseline for detecting misclassified and out-of-distribution examples in neural networks},
  author={Hendrycks, Dan and Gimpel, Kevin},
  journal={arXiv preprint arXiv:1610.02136},
  year={2016}
}

@article{liu2020energy,
  title={Energy-based out-of-distribution detection},
  author={Liu, Weitang and Wang, Xiaoyun and Owens, John and Li, Yixuan},
  journal={Advances in neural information processing systems},
  volume={33},
  pages={21464--21475},
  year={2020}
}

@article{lee2018simple,
  title={A simple unified framework for detecting out-of-distribution samples and adversarial attacks},
  author={Lee, Kimin and Lee, Kibok and Lee, Honglak and Shin, Jinwoo},
  journal={Advances in neural information processing systems},
  volume={31},
  year={2018}
}

@inproceedings{liu2023gen,
  title={Gen: Pushing the limits of softmax-based out-of-distribution detection},
  author={Liu, Xixi and Lochman, Yaroslava and Zach, Christopher},
  booktitle={Proceedings of the IEEE/CVF conference on computer vision and pattern recognition},
  pages={23946--23955},
  year={2023}
}

@inproceedings{deng2009imagenet,
  title={Imagenet: A large-scale hierarchical image database},
  author={Deng, Jia and Dong, Wei and Socher, Richard and Li, Li-Jia and Li, Kai and Fei-Fei, Li},
  booktitle={2009 IEEE conference on computer vision and pattern recognition},
  pages={248--255},
  year={2009},
  organization={Ieee}
}

@inproceedings{ruff2018deep,
  title={Deep one-class classification},
  author={Ruff, Lukas and Vandermeulen, Robert and Goernitz, Nico and Deecke, Lucas and Siddiqui, Shoaib Ahmed and Binder, Alexander and M{\"u}ller, Emmanuel and Kloft, Marius},
  booktitle={International conference on machine learning},
  pages={4393--4402},
  year={2018},
  organization={PMLR}
}

@inproceedings{livernoche2024diffusion,
  title={On Diffusion Modeling for Anomaly Detection},
  author={Livernoche, Victor and Jain, Vineet and Hezaveh, Yashar and Ravanbakhsh, Siamak},
  booktitle={The Twelfth International Conference on Learning Representations},
  pages={7999},
  year={2024}
}

@inproceedings{liu2021rca,
  title={Rca: A deep collaborative autoencoder approach for anomaly detection},
  author={Liu, Boyang and Wang, Ding and Lin, Kaixiang and Tan, Pang-Ning and Zhou, Jiayu},
  booktitle={IJCAI: proceedings of the conference},
  volume={2021},
  pages={1505},
  year={2021}
}

@article{wang2019unsupervised,
  title={Unsupervised representation learning by predicting random distances},
  author={Wang, Hu and Pang, Guansong and Shen, Chunhua and Ma, Congbo},
  journal={arXiv preprint arXiv:1912.12186},
  year={2019}
}

@inproceedings{he2016deep,
  title={Deep residual learning for image recognition},
  author={He, Kaiming and Zhang, Xiangyu and Ren, Shaoqing and Sun, Jian},
  booktitle={Proceedings of the IEEE conference on computer vision and pattern recognition},
  pages={770--778},
  year={2016}
}

@article{dosovitskiy2020image,
  title={An image is worth 16x16 words: Transformers for image recognition at scale},
  author={Dosovitskiy, Alexey and Beyer, Lucas and Kolesnikov, Alexander and Weissenborn, Dirk and Zhai, Xiaohua and Unterthiner, Thomas and Dehghani, Mostafa and Minderer, Matthias and Heigold, Georg and Gelly, Sylvain and others},
  journal={arXiv preprint arXiv:2010.11929},
  year={2020}
}

@inproceedings{devlin2019bert,
  title={Bert: Pre-training of deep bidirectional transformers for language understanding},
  author={Devlin, Jacob and Chang, Ming-Wei and Lee, Kenton and Toutanova, Kristina},
  booktitle={Proceedings of the 2019 conference of the North American chapter of the association for computational linguistics: human language technologies, volume 1 (long and short papers)},
  pages={4171--4186},
  year={2019}
}

@article{liu2019roberta,
  title={Roberta: A robustly optimized bert pretraining approach},
  author={Liu, Yinhan and Ott, Myle and Goyal, Naman and Du, Jingfei and Joshi, Mandar and Chen, Danqi and Levy, Omer and Lewis, Mike and Zettlemoyer, Luke and Stoyanov, Veselin},
  journal={arXiv preprint arXiv:1907.11692},
  year={2019}
}

@inproceedings{yang2025measuring,
  title={Measuring data diversity for instruction tuning: A systematic analysis and a reliable metric},
  author={Yang, Yuming and Nan, Yang and Ye, Junjie and Dou, Shihan and Wang, Xiao and Li, Shuo and Lv, Huijie and Gui, Tao and Zhang, Qi and Huang, Xuan-Jing},
  booktitle={Proceedings of the 63rd Annual Meeting of the Association for Computational Linguistics (Volume 1: Long Papers)},
  pages={18530--18549},
  year={2025}
}

@article{pasarkar2026vendi,
  title={Vendi Novelty Scores for Out-of-Distribution Detection},
  author={Pasarkar, Amey P and Dieng, Adji Bousso},
  journal={arXiv preprint arXiv:2602.10062},
  year={2026}
}

@article{pedregosa2011scikit,
  title={Scikit-learn: Machine learning in Python},
  author={Pedregosa, Fabian and Varoquaux, Ga{\"e}l and Gramfort, Alexandre and Michel, Vincent and Thirion, Bertrand and Grisel, Olivier and Blondel, Mathieu and Prettenhofer, Peter and Weiss, Ron and Dubourg, Vincent and others},
  journal={the Journal of machine Learning research},
  volume={12},
  pages={2825--2830},
  year={2011},
  publisher={JMLR. org}
}

@inproceedings{ahmed2017anomaly,
  title={Anomaly detection on big data in financial markets},
  author={Ahmed, Mohiuddin and Choudhury, Nazim and Uddin, Shahadat},
  booktitle={Proceedings of the 2017 IEEE/ACM International Conference on Advances in Social Networks Analysis and Mining 2017},
  pages={998--1001},
  year={2017}
}

@article{nassif2021machine,
  title={Machine learning for anomaly detection: A systematic review},
  author={Nassif, Ali Bou and Talib, Manar Abu and Nasir, Qassim and Dakalbab, Fatima Mohamad},
  journal={Ieee Access},
  volume={9},
  pages={78658--78700},
  year={2021},
  publisher={Ieee}
}

@article{pang2021deep,
  title={Deep learning for anomaly detection: A review},
  author={Pang, Guansong and Shen, Chunhua and Cao, Longbing and Hengel, Anton Van Den},
  journal={ACM computing surveys (CSUR)},
  volume={54},
  number={2},
  pages={1--38},
  year={2021},
  publisher={ACM New York, NY, USA}
}

@inproceedings{reiss2021panda,
  title={Panda: Adapting pretrained features for anomaly detection and segmentation},
  author={Reiss, Tal and Cohen, Niv and Bergman, Liron and Hoshen, Yedid},
  booktitle={Proceedings of the IEEE/CVF conference on computer vision and pattern recognition},
  pages={2806--2814},
  year={2021}
}

@article{cao2025tad,
  title={Tad-bench: A comprehensive benchmark for embedding-based text anomaly detection},
  author={Cao, Yang and Yang, Sikun and Li, Chen and Xiang, Haolong and Qi, Lianyong and Liu, Bo and Li, Rongsheng and Liu, Ming},
  journal={arXiv preprint arXiv:2501.11960},
  year={2025}
}

@article{vargaftik2021rade,
  title={RADE: resource-efficient supervised anomaly detection using decision tree-based ensemble methods},
  author={Vargaftik, Shay and Keslassy, Isaac and Orda, Ariel and Ben-Itzhak, Yaniv},
  journal={Machine Learning},
  volume={110},
  number={10},
  pages={2835--2866},
  year={2021},
  publisher={Springer}
}

@article{ma2016supervised,
  title={Supervised anomaly detection in uncertain pseudoperiodic data streams},
  author={Ma, Jiangang and Sun, Le and Wang, Hua and Zhang, Yanchun and Aickelin, Uwe},
  journal={ACM Transactions on Internet Technology (TOIT)},
  volume={16},
  number={1},
  pages={1--20},
  year={2016},
  publisher={ACM New York, NY, USA}
}

@inproceedings{goodge2022lunar,
  title={Lunar: Unifying local outlier detection methods via graph neural networks},
  author={Goodge, Adam and Hooi, Bryan and Ng, See-Kiong and Ng, Wee Siong},
  booktitle={Proceedings of the AAAI conference on artificial intelligence},
  volume={36},
  number={6},
  pages={6737--6745},
  year={2022}
}

@inproceedings{guha2016robust,
  title={Robust random cut forest based anomaly detection on streams},
  author={Guha, Sudipto and Mishra, Nina and Roy, Gourav and Schrijvers, Okke},
  booktitle={International conference on machine learning},
  pages={2712--2721},
  year={2016},
  organization={PMLR}
}

@inproceedings{bandaragoda2014efficient,
  title={Efficient anomaly detection by isolation using nearest neighbour ensemble},
  author={Bandaragoda, Tharindu R and Ting, Kai Ming and Albrecht, David and Liu, Fei Tony and Wells, Jonathan R},
  booktitle={2014 IEEE International conference on data mining workshop},
  pages={698--705},
  year={2014},
  organization={IEEE}
}

@article{stein2023exposing,
  title={Exposing flaws of generative model evaluation metrics and their unfair treatment of diffusion models},
  author={Stein, George and Cresswell, Jesse and Hosseinzadeh, Rasa and Sui, Yi and Ross, Brendan and Villecroze, Valentin and Liu, Zhaoyan and Caterini, Anthony L and Taylor, Eric and Loaiza-Ganem, Gabriel},
  journal={Advances in Neural Information Processing Systems},
  volume={36},
  pages={3732--3784},
  year={2023}
}

@article{neshat2025smart,
  title={Smart buildings energy consumption forecasting using adaptive evolutionary bagging extra tree learning models},
  author={Neshat, Mehdi and Thilakaratne, Menasha and El-Abd, Mohammed and Mirjalili, Seyedali and Gandomi, Amir H and Boland, John},
  journal={Energy},
  pages={137130},
  year={2025},
  publisher={Elsevier}
}

@article{zamzmi2025scorecard,
  title={Scorecard for synthetic medical data evaluation},
  author={Zamzmi, Ghada and Subbaswamy, Adarsh and Sizikova, Elena and Margerrison, Edward and Delfino, Jana G and Badano, Aldo},
  journal={Communications Engineering},
  volume={4},
  number={1},
  pages={130},
  year={2025},
  publisher={Nature Publishing Group UK London}
}

@article{denadel2025evaluating,
  title={Evaluating the role of pre-training dataset size and diversity on single-cell foundation model performance},
  author={DenAdel, Alan and Hughes, Madeline and Thoutam, Akshaya and Gupta, Anay and Navia, Andrew W and Fusi, Nicolo and Raghavan, Srivatsan and Winter, Peter S and Amini, Ava P and Crawford, Lorin},
  journal={bioRxiv},
  pages={2024--12},
  year={2025}
}

@article{nielsen2025applications,
  title={Applications of the Vendi score in genomic epidemiology},
  author={Nielsen, Bjarke Frost and Pasarkar, Amey P and Yang, Qiqi and Grenfell, Bryan T and Dieng, Adji Bousso},
  journal={arXiv preprint arXiv:2509.22520},
  year={2025}
}

@article{nguyen2024quality,
  title={Quality-weighted vendi scores and their application to diverse experimental design},
  author={Nguyen, Quan and Dieng, Adji Bousso},
  journal={arXiv preprint arXiv:2405.02449},
  year={2024}
}

@inproceedings{sun2022out,
  title={Out-of-distribution detection with deep nearest neighbors},
  author={Sun, Yiyou and Ming, Yifei and Zhu, Xiaojin and Li, Yixuan},
  booktitle={International conference on machine learning},
  pages={20827--20840},
  year={2022},
  organization={PMLR}
}

@inproceedings{rippel2021modeling,
  title={Modeling the distribution of normal data in pre-trained deep features for anomaly detection},
  author={Rippel, Oliver and Mertens, Patrick and Merhof, Dorit},
  booktitle={2020 25th International Conference on Pattern Recognition (ICPR)},
  pages={6726--6733},
  year={2021},
  organization={IEEE}
}

@misc{simeoni2025dinov3,
  title={{DINOv3}},
  author={Sim{\'e}oni, Oriane and Vo, Huy V. and Seitzer, Maximilian and Baldassarre, Federico and Oquab, Maxime and Jose, Cijo and Khalidov, Vasil and Szafraniec, Marc and Yi, Seungeun and Ramamonjisoa, Micha{\"e}l and Massa, Francisco and Haziza, Daniel and Wehrstedt, Luca and Wang, Jianyuan and Darcet, Timoth{\'e}e and Moutakanni, Th{\'e}o and Sentana, Leonel and Roberts, Claire and Vedaldi, Andrea and Tolan, Jamie and Brandt, John and Couprie, Camille and Mairal, Julien and J{\'e}gou, Herv{\'e} and Labatut, Patrick and Bojanowski, Piotr},
  year={2025},
  eprint={2508.10104},
  archivePrefix={arXiv},
  primaryClass={cs.CV},
  url={https://arxiv.org/abs/2508.10104},
}

@inproceedings{wang2022vim,
  title={Vim: Out-of-distribution with virtual-logit matching},
  author={Wang, Haoqi and Li, Zhizhong and Feng, Litong and Zhang, Wayne},
  booktitle={Proceedings of the IEEE/CVF conference on computer vision and pattern recognition},
  pages={4921--4930},
  year={2022}
}

@article{barcelo2024avoiding,
  title={Avoiding mode collapse in diffusion models fine-tuned with reinforcement learning},
  author={Barcel{\'o}, Roberto and Alc{\'a}zar, Crist{\'o}bal and Tobar, Felipe},
  journal={arXiv preprint arXiv:2410.08315},
  year={2024}
}

@inproceedings{tom2026beyond,
  title={Beyond Reward Maximization: Evaluating the Diversity of Trajectories in Reinforcement Learning with Temporal Vendi Score},
  author={Tom, Stanic and Jiralerspong, Marco and Xiaofeng, Zhang and Vucetic, Danilo and Gidel, Gauthier},
  booktitle={ICLR 2026 Workshop on Lifelong Agents: Learning, Aligning, Evolving}
}

@article{krizhevsky2009learning,
  title={Learning multiple layers of features from tiny images},
  author={Krizhevsky, Alex and Hinton, Geoffrey and others},
  year={2009},
  publisher={Toronto, ON, Canada}
}

@inproceedings{netzer2011reading,
  title={Reading digits in natural images with unsupervised feature learning},
  author={Netzer, Yuval and Wang, Tao and Coates, Adam and Bissacco, Alessandro and Wu, Baolin and Ng, Andrew Y and others},
  booktitle={NIPS workshop on deep learning and unsupervised feature learning},
  volume={2011},
  number={2},
  pages={4},
  year={2011},
  organization={Granada}
}

@article{mu2019mnist,
  title={Mnist-c: A robustness benchmark for computer vision},
  author={Mu, Norman and Gilmer, Justin},
  journal={arXiv preprint arXiv:1906.02337},
  year={2019}
}

@inproceedings{bergmann2019mvtec,
  title={MVTec AD--A comprehensive real-world dataset for unsupervised anomaly detection},
  author={Bergmann, Paul and Fauser, Michael and Sattlegger, David and Steger, Carsten},
  booktitle={Proceedings of the IEEE/CVF conference on computer vision and pattern recognition},
  pages={9592--9600},
  year={2019}
}

@inproceedings{he2016ups,
  title={Ups and downs: Modeling the visual evolution of fashion trends with one-class collaborative filtering},
  author={He, Ruining and McAuley, Julian},
  booktitle={proceedings of the 25th international conference on world wide web},
  pages={507--517},
  year={2016}
}

@article{zhang2015character,
  title={Character-level convolutional networks for text classification},
  author={Zhang, Xiang and Zhao, Junbo and LeCun, Yann},
  journal={Advances in neural information processing systems},
  volume={28},
  year={2015}
}

@inproceedings{askari2024improving,
  title={Improving geo-diversity of generated images with contextualized vendi score guidance},
  author={Askari Hemmat, Reyhane and Hall, Melissa and Sun, Alicia and Ross, Candace and Drozdzal, Michal and Romero-Soriano, Adriana},
  booktitle={European Conference on Computer Vision},
  pages={213--229},
  year={2024},
  organization={Springer}
}

@article{dieng2025unified,
  title={A Unified and Predictive Measure of Functional Diversity},
  author={Dieng, Adji Bousso and Pasarkar, Amey},
  journal={arXiv preprint arXiv:2509.16133},
  year={2025}
}

@article{hu2026evaluating,
  title={Evaluating the Diversity of AI-Generated Content with Diversity Profiles},
  author={Hu, Xiuyuan and Hou, Xuege and Liu, Guoqing and Zhao, Yang and Li, Jieran and Sun, Dongbiao and Hern{\'a}ndez-Lobato, Jos{\'e} Miguel and Zhang, Hao and Liu, Xue},
  journal={arXiv preprint arXiv:2608.17731},
  year={2026}
}

\newpage 
\appendix

\section{Proof of Proposition \ref{prop:first-order-vas}}
\label{sec:proof}

Removing $x_i$ induces the perturbation
\[
\Delta_i := \rho_{-i} - \rho
=
-\frac{x_i x_i^\top - \rho}{N-1}.
\]
Moreover,
\[
\|\Delta_i\|
\leq
\frac{\|x_ix_i^\top\|+\|\rho\|}{N-1}
\leq
\frac{2}{N-1} =  O\!\left(\frac{1}{N}\right),
\]
because $\|x_i\|_2=1$ and $\|\rho\|\leq \operatorname{Tr}(\rho)=1$.

We now compute the first variation of $\log \mathrm{VS}_q$.

\paragraph{Case $q \neq 1$.}
\[
\log \mathrm{VS}_q(\rho)
=
\frac{1}{1-q} \log \operatorname{Tr}(\rho^q).
\]
Here $D \log \mathrm{VS}_q(\rho)[\Delta]$ denotes the directional derivative of $\log \mathrm{VS}_q$ at $\rho$ in the perturbation direction $\Delta$. Then,
\[
D \log \mathrm{VS}_q(\rho)[\Delta]
=
\frac{q}{(1-q)\operatorname{Tr}(\rho^q)}
\operatorname{Tr}(\rho^{q-1}\Delta)
=
\operatorname{Tr}(G_q \Delta).
\]

\paragraph{Case $q = 1$.}
\[
\log \mathrm{VS}_1(\rho)
=
-\operatorname{Tr}(\rho \log \rho).
\]
Using spectral calculus,
\[
D \log \mathrm{VS}_1(\rho)[\Delta]
=
-\operatorname{Tr}((\log \rho + I)\Delta)
=
\operatorname{Tr}(G_q \Delta).
\]

Thus, in both cases,
\[
D \log \mathrm{VS}_q(\rho)[\Delta]
=
\operatorname{Tr}(G_q \Delta).
\]

By Taylor's theorem,
\[
\log \mathrm{VS}_q(\rho + \Delta_i)
=
\log \mathrm{VS}_q(\rho)
+
\operatorname{Tr}(G_q \Delta_i)
+
O(\|\Delta_i\|^2).
\]
Substituting $\Delta_i$ gives
\[
\log \mathrm{VS}_q(\rho_{-i}) - \log \mathrm{VS}_q(\rho)
=
-\frac{1}{N-1}
\operatorname{Tr}\!\left(G_q (x_i x_i^\top - \rho)\right)
+
O\!\left(\frac{1}{N^2}\right).
\]

Using $\operatorname{Tr}(G_q x_i x_i^\top) = x_i^\top G_q x_i$, we obtain
\[
\log \mathrm{VS}_q(\rho) - \log \mathrm{VS}_q(\rho_{-i})
=
\frac{x_i^\top G_q x_i - \operatorname{Tr}(G_q \rho)}{N-1}
+
O\!\left(\frac{1}{N^2}\right),
\]
which completes the proof.

\begin{figure}[t]
    \centering
    \includegraphics[width=0.95\linewidth]{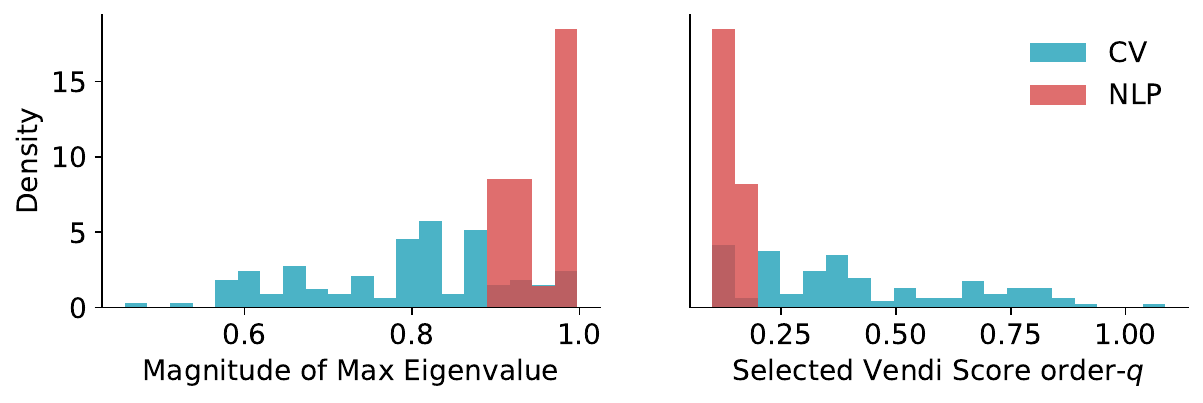}
    \caption{Distribution of principal eigenvalue (left) and the selected $q$ orders (right) on all tested ADBENCH benchmarks. The geometry of CV datasets is significantly different than those from NLP datasets, so we select different $q$ orders to account for this.}
    \label{fig:q_histogram}
\end{figure}

\section{Comparison of VAS to Alternative Spectral Scores}
\label{sec:spectral_comparison}
VAS belongs to a broader family of spectral quadratic anomaly
scores. For $\ell_2$-normalized embeddings
$x_1,\ldots,x_N\in\mathbb{R}^D$, consider the eigendecomposition
\begin{equation}
\rho = \frac{1}{N}\sum_{i=1}^N x_i x_i^\top
= U\operatorname{diag}(\lambda_1,\ldots,\lambda_D)U^\top,
\end{equation}
where $\lambda_1\geq\cdots\geq\lambda_D\geq 0$.
Writing $z_j=u_j^\top x$, any spectral function
$M=f(\rho)$ defines a quadratic score
\begin{equation}
s_M(x)=x^\top Mx
=\sum_j f(\lambda_j)z_j^2.
\end{equation}
Different spectral detectors differ in how the choice of $M$, and therefore how they weight the eigen-directions of $\rho$. From Proposition 3.1,
VAS assigns the weights
\begin{equation}
f_q(\lambda_j)=
\begin{cases}
\displaystyle
\frac{q\lambda_j^{q-1}}
{(1-q)\operatorname{Tr}(\rho^q)}, & q\neq 1,\\[0.8em]
-\log\lambda_j-1, & q=1.
\end{cases}
\label{eq:vas_spectral_weights}
\end{equation}
As in the main text, spectral functions are evaluated on the
positive eigenspace of $\rho$. For $0<q<1$, VAS assigns
larger positive weights to smaller eigenvalues. For $q>1$,
the weights are negative, with larger magnitudes on leading
eigen-directions. Thus, the order $q$ controls how spectral
directions contribute to diversity influence.

\paragraph{Mahalanobis distance.}
For mean-centered features, the covariance equals $\rho$,
and the squared Mahalanobis score is
\begin{equation}
s_{\mathrm{Mah}}(x)
=x^\top\rho^\dagger x
=\sum_{\lambda_j>0}\lambda_j^{-1}z_j^2,
\end{equation}
where $\rho^\dagger$ is the pseudoinverse.
As $q\to 0^+$, the VAS spectral weights become
proportional to $\lambda_j^{-1}$. More precisely,
\begin{equation}
\lim_{q\to 0^+}
\frac{(1-q)\operatorname{Tr}(\rho^q)}{q}
s_{\mathrm{VAS},q}(x)
=x^\top\rho^\dagger x.
\end{equation}
Thus, VAS approaches the Mahalanobis ranking in this
limit. Unlike standard Mahalanobis distance, however,
VAS operates on the uncentered second-moment matrix
and adaptively selects $q$ rather than fixing inverse
eigenvalue weights.

\paragraph{PCA reconstruction error.}
Let $U_r$ contain the leading $r$ eigenvectors of $\rho$.
The corresponding PCA residual score is
\begin{equation}
s_{\mathrm{PCA},r}(x)
=\|(I-U_rU_r^\top)x\|_2^2
=\sum_{j>r}z_j^2.
\end{equation}
PCA therefore assigns zero weight to the leading $r$
directions and equal weight to the remaining directions.
Kernel PCA applies an analogous reconstruction principle
in a kernel feature space. In contrast, VAS uses a smooth
spectral weighting without imposing a hard rank cutoff.

\paragraph{Global similarity.}
For $q=2$, the VAS score simplifies to
\begin{equation}
\begin{aligned}
s_{\mathrm{VAS},2}(x_i)
&=-\frac{2}{\operatorname{Tr}(\rho^2)}
x_i^\top\rho x_i\\
&=-\frac{2}{N\operatorname{Tr}(\rho^2)}
\sum_{j=1}^N(x_i^\top x_j)^2.
\end{aligned}
\end{equation}
Thus, at $q=2$, VAS is ranking-equivalent to negative
average squared cosine similarity, favoring samples
that are globally dissimilar to the dataset.

VAS is related to classical spectral detectors, but its weighting arises from the leave-one-out influence of a sample on dataset diversity. Rather than prescribing a fixed spectral filter, VAS adapts its weighting to the dataset through the Vendi Score order $q$.

\section{Proof of Proposition \ref{prop:vas_component_spectral_density}} \label{sec:vas_component_spectral_density}
Since \(P=\sum_{c=1}^K \pi_c P_c\), we have
\[
\rho
=
\mathbb E_{x\sim P}[xx^\top]
=
\sum_{c=1}^K \pi_c \mathbb E_{x\sim P_c}[xx^\top]
=
\sum_{c=1}^K \pi_c\rho_c.
\]
Because each component \(P_c\) is supported on subspace \(U_c\), the operator
\(\rho_c\) is supported on \(U_c\). Since the subspaces are mutually
orthogonal, \(\rho\) is block diagonal with block \(\pi_c\rho_c\) on
\(U_c\). Therefore, for \(x\in U_c\),
\[
\rho^{q-1}x
=
(\pi_c\rho_c)^{q-1}x
=
\pi_c^{q-1}\rho_c^{q-1}x.
\]
Substituting into the definition of \(G_q\) gives
\[
x^\top G_qx
=
\frac{q}{(1-q)\operatorname{Tr}(\rho^q)}
x^\top \rho^{q-1}x
=
\frac{q}{(1-q)\operatorname{Tr}(\rho^q)}
\pi_c^{q-1}x^\top \rho_c^{q-1}x.
\]

Taking expectation over \(x\sim P_c\),
\[
\mathbb E_{x\sim P_c}[x^\top G_qx]
=
\frac{q}{(1-q)\operatorname{Tr}(\rho^q)}
\pi_c^{q-1}
\mathbb E_{x\sim P_c}[x^\top \rho_c^{q-1}x].
\]
Using \(\mathbb E_{x\sim P_c}[xx^\top]=\rho_c\),
\[
\mathbb E_{x\sim P_c}[x^\top \rho_c^{q-1}x]
=
\operatorname{Tr}\!\left(\rho_c^{q-1}\mathbb E_{x\sim P_c}[xx^\top]\right)
=
\operatorname{Tr}(\rho_c^q).
\]
Finally, since
\[
\mathrm{VS}_q(\rho_c)
=
\left(\operatorname{Tr}(\rho_c^q)\right)^{1/(1-q)},
\]
we have
\[
\operatorname{Tr}(\rho_c^q)
=
\mathrm{VS}_q(\rho_c)^{1-q}.
\]
Thus
\[
\pi_c^{q-1}\operatorname{Tr}(\rho_c^q)
=
\pi_c^{q-1} \mathrm{VS}_q(\rho_c)^{1-q}
=
\left(\frac{\pi_c}{\mathrm{VS}_q(\rho_c)}\right)^{q-1},
\]
which proves the claim.

\section{ADBENCH Embedding Datasets}
\label{sec:datasest}
Here we summarize the ADBENCH embedding datasets. For all image datasets, we use embeddings from both ResNet18 \citep{he2016deep} and ViT models \citep{dosovitskiy2020image}. For text datasets, we use embeddings from BERT \citep{devlin2019bert} and RoBERTa \citep{liu2019roberta}. 

\begin{itemize}
    \item \textbf{CIFAR-10 (10 splits):} Each split treats one class as normal and the remaining nine classes as anomalies, which are downsampled to 5\% of the total dataset. Each split contains 5{,}263 samples. Results are averaged across all 10 choices of the normal class. Evaluated using both 512- and 1000-dimensional image embeddings.

    \item \textbf{FashionMNIST (10 splits):} Same protocol as CIFAR-10, with one class designated as normal and others downsampled to 5\% anomalies. Each split contains 6,315 samples. Results are averaged across splits. Evaluated with 512- and 1000-dimensional embeddings.

    \item \textbf{SVHN (10 splits):} Same one-class protocol as CIFAR-10. Each split corresponds to a different digit as the normal class, with remaining digits downsampled to 5\% anomalies. Sample sizes range from 4,904 to 10,000. Evaluated with 512- and 1000-dimensional embeddings.

    \item \textbf{MNIST-C (16 splits):} Corrupted variants of MNIST, where original MNIST images are treated as normal and corrupted versions (e.g., noise, blur, etc.) are treated as anomalies. Each split corresponds to a different corruption type and contains 10{,}000 samples. Evaluated with 512- and 1000-dimensional embeddings.

    \item \textbf{MVTec-AD (15 splits):} Industrial anomaly detection dataset consisting of images of manufactured objects with defects. Each object category (15 total) defines a separate anomaly detection task, where defect-free images are normal and defective images are anomalous. Sample sizes range from 102 to 501 per category. Evaluated with 512- and 1000-dimensional embeddings.

    \item \textbf{20 Newsgroups (6 splits):} For each split, one topic group is treated as normal and the remaining groups are treated as anomalies. The six normal classes are \textit{computer}, \textit{recreation}, \textit{science}, \textit{miscellaneous}, \textit{politics}, and \textit{religion}. Sample sizes are 615, 1{,}532, 1{,}657, 2{,}497, 2{,}514, and 3{,}090. Evaluated using 768-dimensional text embeddings.

    \item \textbf{AG News (4 splits):} Multi-class news dataset where each split treats one class as normal and the remaining classes are downsampled to 5\% anomalies. Each split contains 10{,}000 samples. Evaluated with 768-dimensional embeddings.

    \item \textbf{Amazon Reviews (1 dataset):} Binary sentiment dataset with 10,000 samples. Negative reviews are treated as anomalies and positive reviews as normal. Evaluated using 768-dimensional embeddings.

    \item \textbf{IMDB (1 dataset):} Binary sentiment dataset with 10,000 samples. Negative reviews are treated as anomalies and positive reviews as normal. Evaluated using 768-dimensional embeddings.

    \item \textbf{Yelp (1 dataset):} Review dataset with 10,000 samples. Reviews with 0--1 stars are treated as anomalies, while 3--4 star reviews are treated as normal. Evaluated using 768-dimensional embeddings.
\end{itemize}

\begin{table}[t]
\centering
\small
\setlength{\tabcolsep}{4.5pt}
\renewcommand{\arraystretch}{1.15}
\resizebox{\linewidth}{!}{%
\begin{tabular}{llcccc}
\toprule
\textbf{Dataset} & \textbf{Encoder}
& \textbf{Approx. AUROC}
& \textbf{Exact AUROC}
& $\boldsymbol{\Delta}$ \textbf{AUROC (\%)}
& \textbf{Spearman } $\boldsymbol{\rho}$ \\
\midrule
CIFAR-10 & ResNet18 & 0.745 & 0.745 & -0.01 & 1.000 \\
CIFAR-10 & ViT & 0.941 & 0.941 & -0.02 & 1.000 \\
FashionMNIST & ResNet18 & 0.911 & 0.911 & -0.02 & 1.000 \\
FashionMNIST & ViT & 0.930 & 0.929 & 0.00 & 1.000 \\
MNIST-C & ResNet18 & 0.793 & 0.793 & 0.00 & 1.000 \\
MNIST-C & ViT & 0.739 & 0.739 & 0.01 & 1.000 \\
MVTec-AD & ResNet18 & 0.799 & 0.796 & -0.41 & 0.990 \\
MVTec-AD & ViT & 0.896 & 0.906 & 1.16 & 0.985 \\
SVHN & ResNet18 & 0.657 & 0.657 & 0.00 & 1.000 \\
SVHN & ViT & 0.579 & 0.579 & 0.06 & 1.000 \\
\midrule
20News & BERT & 0.626 & 0.630 & 0.48 & 0.997 \\
20News & RoBERTa & 0.749 & 0.748 & -0.08 & 0.995 \\
AG News & BERT & 0.676 & 0.676 & 0.03 & 1.000 \\
AG News & RoBERTa & 0.881 & 0.881 & 0.01 & 1.000 \\
Amazon & BERT & 0.608 & 0.608 & 0.00 & 1.000 \\
Amazon & RoBERTa & 0.544 & 0.544 & 0.00 & 1.000 \\
IMDB & BERT & 0.502 & 0.502 & -0.01 & 1.000 \\
IMDB & RoBERTa & 0.530 & 0.530 & 0.00 & 1.000 \\
Yelp & BERT & 0.679 & 0.679 & 0.00 & 1.000 \\
Yelp & RoBERTa & 0.566 & 0.566 & 0.00 & 1.000 \\
\bottomrule
\end{tabular}%
}
\caption{
Comparison of exact and approximate VAS across
148 ADBENCH benchmark instances.
Results are averaged across dataset splits for each
dataset--encoder combination.
$\Delta$ AUROC (\%) is the percentage change from
approximate to exact VAS.
Spearman's $\rho$ measures the rank correlation between
exact and approximate anomaly scores.
The approximation preserves both anomaly rankings and
downstream detection performance across datasets.
}
\label{tab:approx_adbench}
\end{table}

\section{Baselines}
\label{sec:baselines}
Here we review the baseline methods we compare against. 
In Figure \ref{fig:latency}, all baselines were run on a single NVIDIA A6000 GPU with 4 Intel Xeon Gold 5320, 2.20GHz CPUs, with results generated over $25$ GPU hours. 
\\\\
\textbf{IForest.} \cite{liu2008isolation} introduced Isolation Forest to take advantage of the idea that anomalies are easier to separate than a normal point from the rest of the dataset. IForest builds an ensemble of isolation trees based on random partitions of the feature space, and detects anomalies based on average how many partitions are needed to isolate a sample. The hyperparameters are the number of samples used to build each tree, and the total number of trees. IForest relies on the feature being easy to partition, which may not be true when features are highly correlated. 

\textbf{DIF.} \cite{xu2023deep} improved on IForest by noting that the linear partitions are not expressive enough for most datasets. The resulting method, Deep Isolation Forest (DIF), augments isolation forests with random deep representations: instead of directly applying axis-parallel splits in the input space, DIF first maps data through an ensemble of randomly initialized neural networks, and then applies standard isolation trees in this transformed space.

DIF is primarily designed for tabular data where nonlinear feature transformations must be learned. On embedding datasets, the representation are already highly expressive and semantically structured. Applying additional random neural projections can distort this geometry, potentially destroying meaningful similarity structure and degrading distance-based anomaly signals. As a result, DIF may offer limited benefit on high-quality embedding datasets, while incurring additional computational overhead.

\textbf{LOF.} \cite{breunig2000lof} designed Local Outlier Factor (LOF) to detect anomalies based on local neighborhoods. In particular, LOF measures how dense a sample's neighborhood is relative to its nearest neighbors. The main assumption is that anomalies will have significantly less dense neighborhoods than their neighbors. This identifies anomalies at a local level, but will ignore global structure, and requires the explicit construction of neighborhoods based on a fixed nearest-neighbor cutoff $k$. LOF requires $O(N^2)$ complexity to construct the nearest-neighbor distances. 

\textbf{DN2.} K-Nearest Neighbors on neural embeddings was explored by \cite{bergman2020deep} as a simple heuristic for anomaly detection. Anomaly scores are computed as the average distance between a test embedding and its $k$ nearest neighbors in the training set. Identifying the nearest neighbors requires $O(N^2)$ time and similar to LOF, requires the construction of explicit local neighborhoods that ignore global structure. 

\textbf{GMD.}
Gaussian feature modeling has been used for anomaly detection in pretrained deep feature spaces \citep{rippel2021modeling}. To   this idea to our fully unsupervised setting, where the unlabeled data may be multi-modal and contaminated, we use a Gaussian-mixture Mahalanobis distance baseline. We fit a GMM with a tied covariance matrix $\Sigma$, select the number of components
$K$ by BIC, assign each sample $x$ to its most likely component with mean $\mu_k$, and score it by
\[
d(x) = (x - \mu_k)^\top \Sigma^{-1} (x - \mu_k),
\]
with larger values indicating greater deviation from the learned feature distribution. The main computational cost is EM fitting, which scales as $O(TKND^2)$ for $T$ iterations, $K$ components, $N$ samples, and dimension $D$. Following recommendations for Mahalanobis scoring in deep feature spaces \citep{mueller2025mahalanobis++}, we apply GMD to $\ell_2$-normalized features.

\textbf{SLAD.} Scale Learning for Anomaly Detection (SLAD) is a deep learning approach for identifying anomalies \citep{xu2023fascinating}. It learns representations by introducing auxiliary supervisory signals derived from scale consistency, encouraging the model to produce embeddings that are invariant across randomly sampled feature subspaces. SLAD constructs multiple random subspaces of the input features and trains a neural network to predict consistent representations across these subspaces, thereby capturing stable patterns of normal data while amplifying deviations for anomalies. To improve robustness, SLAD uses an ensemble of subspace learners, where each model operates on different sampled feature subsets and their outputs are aggregated into a final anomaly score. 

\textbf{ADERH.} Adaptive Density Estimation via Random Hyperspheres (ADERH) is an unsupervised anomaly detection method that estimates local data density by constructing an ensemble of random hyperspheres in feature space \citep{durani2025anomaly}. Each hypersphere is defined by a randomly sampled center and radius, and its contribution is weighted based on how many training points fall inside it. Regions that consistently contain many points across sampled hyperspheres are considered high-density (normal), while points that fall into few or low-density hyperspheres are assigned higher anomaly scores. This ensemble-based strategy allows ADERH to approximate complex density structure without explicitly fitting a parametric model, and aligns with the general principle that anomalies correspond to low-density regions in the data. ADERH's complexity is $O(N \cdot M \cdot D)$, where $N$ is the number of samples, $M$ is the number of hyperspheres, and $D$ is the feature dimension.

ADERH implicitly assumes that normal data can be well-approximated by a union of compact, spherical regions that are separable from anomalies. While this assumption is often reasonable for tabular data, it breaks down in modern embedding spaces. Embeddings tend to lie on continuous, dense manifolds rather than forming isolated clusters \citep{ethayarajh2019contextual}. As a result, while ADERH can perform strongly on structured tabular datasets, it may be less appropriate for anomaly detection on modern embedding representations.

\textbf{LODA.} Lightweight On-line Detector of Anomalies (LODA) is an ensemble, training-free approach for anomaly detection based on sparse random projections \citep{pevny2016loda}. Specifically, it projects high-dimensional data into many one-dimensional subspaces, fits simple histogram-based density estimators in each projection, and aggregates these to estimate the overall likelihood of a sample. LODA is quite fast: the runtime scales with $O(NMD)$ for building $M$ projections and histograms in $D$ dimensions. 

LODA assumes that meaningful structure can be captured through the random projections and corresponding densities. While this may be reasonable for tabular data, it breaks down in embedding spaces where information is distributed across many correlated dimensions and lies on a structured manifold. Random projections to 1D can destroy this geometry, making density estimates unreliable and reducing anomaly separability. 

\textbf{KPCA.} \cite{fang2025kernel} demonstrated that Kernel PCA reconstruction errors can be a useful signal for OOD detection. Here, we apply their approach to anomaly detection. The KPCA approach uses a fused cosine-gaussian kernel to fit the data and compute reconstruction errors. \cite{fang2025kernel} showed that these errors can be computed efficiently for OOD detection using Random Fourier Features (RFFs). In all, this approach has time complexity $O(NDm)$ for an $N$ sample dataset in $D$ dimensions approximated with a $m$-dimensional RFF. 

VAS and KPCA both rely on kernels, but use different motivations for anomaly detection. KPCA learns a low-rank approximation of the data and scores anomalies based on reconstruction error, implicitly assuming that normal data lies near a low-dimensional manifold. In contrast, VAS measures how each sample perturbs the eigenvalue spectrum of the similarity matrix. As a result, KPCA is sensitive primarily to deviations from the leading eigenvalues, whereas VAS captures changes to the full spectral distribution, including effects in the tail. This allows VAS to detect anomalies that may lie within the dominant subspace but still alter the overall diversity structure of the dataset.

\textbf{LUNAR.} LUNAR is a graph-neural-network-based local anomaly detector that learns to score samples from their nearest-neighbor graph \citep{goodge2022lunar}. It generalizes
classical local outlier methods by replacing fixed neighborhood statistics with trainable message passing over local neighborhoods. Unlike fixed neighborhood scores, LUNAR incurs additional cost from constructing the nearest-neighbor graph and training a GNN, but can learn a more adaptive local anomaly score once the graph is built.

\textbf{DTE.} Diffusion Time Estimation (DTE) identifies anomalies based on how diffuse they are from the normal data distribution \citep{livernoche2024diffusion}. DTE estimates the posterior distribution over diffusion time for a given sample, which serves as a proxy for its distance from the data manifold. Intuitively, normal samples lie close to the data manifold and correspond to small diffusion times, while anomalous samples require larger diffusion times (i.e., more noise) to be explained. A neural network is trained to predict this diffusion time, and the anomaly score is given by the predicted mean or mode.

We focus on the categorical parametrization, which models diffusion time as a discrete distribution over bins. This variant is the empirically strongest.

\begin{table*}[t]
\centering
\scriptsize
\setlength{\tabcolsep}{2.2pt}
\begin{tabular}{llcccccccccccc}
\toprule
Dataset & Encoder & VAS & ADERH & DIF & IForest & LUNAR & DN2 & LODA & LOF & KPCA & SLAD & GMD & DTE \\
\midrule

CIFAR10 & ResNet18
& \third{0.153} & 0.093 & 0.112 & 0.088 & 0.104 & \second{0.163} & 0.091 & 0.122 & \best{0.166} & 0.145 & 0.141 & 0.103 \\

CIFAR10 & ViT
& \second{0.659} & \third{0.541} & 0.326 & 0.360 & 0.389 & 0.386 & 0.365 & 0.072 & 0.486 & \best{0.670} & 0.491 & 0.109 \\

FashionMNIST & ResNet18
& \third{0.449} & 0.370 & 0.395 & 0.325 & 0.319 & 0.337 & 0.321 & 0.128 & \best{0.495} & \second{0.450} & 0.436 & 0.335 \\

FashionMNIST & ViT
& \best{0.546} & \third{0.508} & 0.325 & 0.423 & 0.333 & 0.346 & 0.381 & 0.085 & 0.451 & 0.508 & \second{0.511} & 0.115 \\

MNIST-C & ResNet18
& \second{0.237} & 0.232 & 0.223 & 0.173 & 0.192 & 0.198 & 0.208 & 0.123 & 0.214 & \best{0.256} & \third{0.235} & 0.227 \\

MNIST-C & ViT
& 0.226 & \best{0.269} & 0.220 & \second{0.242} & 0.192 & 0.216 & 0.200 & 0.090 & 0.200 & \third{0.231} & 0.201 & 0.091 \\

MVTec-AD & ResNet18
& \second{0.642} & 0.577 & 0.570 & 0.565 & 0.582 & \best{0.648} & 0.520 & 0.589 & 0.578 & 0.628 & \third{0.637} & 0.543 \\

MVTec-AD & ViT
& \second{0.748} & 0.648 & 0.612 & 0.626 & 0.712 & \third{0.723} & 0.591 & 0.690 & 0.684 & 0.673 & \best{0.750} & 0.489 \\

SVHN & ResNet18
& \third{0.092} & 0.076 & 0.080 & 0.071 & 0.081 & \best{0.095} & 0.076 & 0.085 & 0.091 & 0.091 & \second{0.093} & 0.076 \\

SVHN & ViT
& 0.066 & 0.056 & 0.055 & 0.056 & 0.062 & \second{0.067} & 0.056 & \best{0.068} & 0.063 & 0.060 & \third{0.066} & 0.057 \\

20News & BERT
& \best{0.088} & 0.066 & 0.069 & 0.065 & 0.075 & 0.076 & 0.066 & \third{0.083} & 0.069 & 0.080 & \second{0.084} & 0.069 \\

20News & RoBERTa
& \best{0.148} & 0.080 & 0.074 & 0.076 & 0.112 & \third{0.125} & 0.069 & 0.111 & \second{0.127} & 0.112 & 0.115 & 0.055 \\

Agnews & BERT
& \second{0.094} & 0.066 & 0.065 & 0.064 & \third{0.088} & 0.077 & 0.062 & \best{0.130} & 0.072 & 0.080 & 0.087 & 0.069 \\

Agnews & RoBERTa
& \best{0.334} & 0.125 & 0.089 & 0.089 & 0.199 & \third{0.249} & 0.081 & 0.115 & \second{0.272} & 0.188 & 0.213 & 0.054 \\

Amazon & BERT
& \third{0.063} & 0.060 & 0.059 & 0.058 & \second{0.063} & 0.061 & 0.055 & 0.058 & 0.060 & \best{0.064} & 0.062 & 0.054 \\

Amazon & RoBERTa
& 0.055 & 0.048 & 0.052 & 0.050 & 0.053 & \third{0.055} & 0.052 & \second{0.056} & 0.055 & 0.053 & 0.052 & \best{0.059} \\

Imdb & BERT
& 0.047 & 0.047 & 0.047 & 0.047 & 0.047 & 0.049 & 0.044 & \third{0.050} & \best{0.050} & \second{0.050} & 0.049 & 0.047 \\

Imdb & RoBERTa
& \second{0.056} & 0.046 & 0.053 & 0.050 & 0.053 & 0.054 & 0.053 & \best{0.056} & \third{0.055} & 0.052 & 0.052 & 0.049 \\

Yelp & BERT
& \second{0.086} & 0.074 & 0.074 & 0.070 & \best{0.086} & 0.083 & 0.064 & \third{0.084} & 0.075 & 0.080 & 0.083 & 0.049 \\

Yelp & RoBERTa
& 0.057 & 0.049 & 0.052 & 0.051 & \third{0.061} & \best{0.063} & 0.053 & \second{0.063} & 0.059 & 0.058 & 0.058 & 0.056 \\

\midrule
Average Rank
& & \best{2.90} & 7.90 & 8.45 & 9.65 & 6.00 & \third{4.25} & 9.95 & 6.10 & 4.90 & 4.40 & \second{4.20} & 9.30 \\
p-Value
& & - & \smallscript{1.12e-14} & \smallscript{2.14e-20} & \smallscript{2.79e-20} & \smallscript{4.42e-20} & \smallscript{3.84e-09} & \smallscript{2.42e-21} & \smallscript{1.89e-12} & \smallscript{1.07e-10} & \smallscript{3.59e-08} & \smallscript{6.62e-11} & \smallscript{4.99e-22} \\
\bottomrule
\end{tabular}
\caption{AP across dataset families and encoders. For multi-split datasets, we report average AP. \best{Orange} marks best, \second{blue} marks 2nd. Bottom-row p-values are paired Wilcoxon tests: VAS vs. each baseline over all 148 dataset-split/encoder instances.}
\label{tab:AP_main}
\end{table*}

\section{Grid Search Experiment}
\label{sec:grid-search}

We perform a discrete grid search over key hyperparameters for each method, while keeping all other settings fixed to their benchmark or paper-specified defaults. Below, we list for each method: (i) the default configuration, (ii) the hyperparameters swept over, and (iii) a brief description of each hyperparameter.

\begin{itemize}
    \item \textbf{ADERH:} Default: $n_{\text{estimators}}=256$, subset size $n=18$.  
    Sweep: $n_{\text{estimators}} \in \{100, 200, 300\}$, $n \in \{8, 18, 24\}$.  
    Here, $n_{\text{estimators}}$ is the number of hyperspheres in the ensemble, and $n$ is the number of samples used to define each hypersphere.

    \item \textbf{DIF:} Default: $n_{\text{estimators}}=6$, $n_{\text{ensemble}}=50$.  
    Sweep: $n_{\text{estimators}} \in \{25, 50, 100\}$, $n_{\text{ensemble}} \in \{2, 4, 6\}$.  
    $n_{\text{estimators}}$ is the number of isolation trees, and $n_{\text{ensemble}}$ is the number of random neural representations used.

    \item \textbf{DTE (Categorical):} Default: $\texttt{num\_bins}=7$ (all other training hyperparameters fixed as in \cite{livernoche2024diffusion}).  
    Sweep: $\texttt{num\_bins} \in \{3, 7, 15\}$.  
    $\texttt{num\_bins}$ is the number of discrete diffusion time bins in the categorical parametrization.

    \item \textbf{IForest:} Default: $n_{\text{estimators}}=100$, $\texttt{max\_samples}=256$.  
    Sweep: $n_{\text{estimators}} \in \{50, 100, 200\}$, $\texttt{max\_samples} \in \{128, 256, 512\}$.  
    $n_{\text{estimators}}$ is the number of trees, and $\texttt{max\_samples}$ is the subsample size used to build each tree.

    \item \textbf{KNN:} Default: $n_{\text{neighbors}}=5$ (with internal row-wise $\ell_2$ normalization).  
    Sweep: $n_{\text{neighbors}} \in \{5, 10, 20\}$.  
    $n_{\text{neighbors}}$ is the number of nearest neighbors used to compute the anomaly score.

    \item \textbf{LODA:} Default: $n_{\text{random\_cuts}}=100$, $n_{\text{bins}}=10$.  
    Sweep: $n_{\text{random\_cuts}} \in \{50, 100, 200\}$, $n_{\text{bins}} \in \{5, 10, 25\}$.  
    $n_{\text{random\_cuts}}$ is the number of random projections (histograms), and $n_{\text{bins}}$ is the number of bins per histogram.

    \item \textbf{LOF:} Default: $n_{\text{neighbors}}=20$.  
    Sweep: $n_{\text{neighbors}} \in \{5, 10, 20, 40\}$.  
    $n_{\text{neighbors}}$ is the number of neighbors used to estimate local density.

    \item \textbf{kPCA:} Default: $\gamma=2.5$, $\texttt{rff\_dim}=4096$, $\texttt{exp\_var\_ratio}=0.5$.  
    Sweep: $\gamma \in \{2.0, 2.5, 3.0\}$, $\texttt{exp\_var\_ratio} \in \{0.5, 0.7, 0.9\}$.  
    $\gamma$ controls the kernel bandwidth, $\texttt{rff\_dim}$ is the number of random Fourier features, and $\texttt{exp\_var\_ratio}$ is the fraction of variance retained in PCA.

    \item \textbf{SLAD:} Default: $n_{\text{ensemble}}=20$, $\texttt{subspace\_pool\_size}=50$.  
    Sweep: $n_{\text{ensemble}} \in \{10, 20, 50\}$, $\texttt{subspace\_pool\_size} \in \{25, 50, 100\}$.  
    $n_{\text{ensemble}}$ is the number of subspace learners, and $\texttt{subspace\_pool\_size}$ is the number of candidate feature subspaces.

    \item \textbf{LUNAR:} Default: $n_{\text{neighbors}}=5$, $\texttt{sampling strategy}=\texttt{mixed}$.  
    Sweep: $n_{\text{neighbors}} \in \{5, 10, 20\}$, $\texttt{sampling strategy} \in \{\texttt{subspace}, \texttt{mixed}, \texttt{uniform}\}$.  
    $n_{\text{neighbors}}$ is the number of neighbors used when constructing the KNN graph, and $\texttt{sampling strategy}$ describes how the contrastive samples are constructed.
\end{itemize}

In tables \ref{tab:auroc_sweep} and \ref{tab:ap_sweep}, we report results using the best-performing configuration within the specified grid.

\begin{table*}[t]
\centering
\scriptsize
\setlength{\tabcolsep}{2.2pt}
\begin{tabular}{llcccccccccccc}
\toprule
Dataset & Encoder & VAS & ADERH & DIF & IForest & LUNAR & DN2 & LODA & LOF & KPCA & SLAD & GMD & DTE \\
\midrule

CIFAR10 & ResNet18
& \best{0.745} & 0.632 & 0.631 & 0.633 & 0.663 & \second{0.739} & 0.651 & 0.689 & 0.734 & \third{0.736} & 0.732 & 0.655 \\

CIFAR10 & ViT
& \second{0.941} & \third{0.924} & 0.788 & 0.892 & 0.876 & 0.831 & 0.900 & 0.444 & 0.610 & \best{0.946} & 0.898 & 0.578 \\

FashionMNIST & ResNet18
& \second{0.911} & 0.869 & 0.832 & 0.839 & 0.867 & 0.838 & 0.853 & 0.576 & 0.895 & \third{0.904} & \best{0.912} & 0.823 \\

FashionMNIST & ViT
& \best{0.930} & 0.911 & 0.776 & 0.896 & 0.820 & 0.837 & 0.893 & 0.493 & 0.900 & \third{0.923} & \second{0.925} & 0.656 \\

MNIST-C & ResNet18
& \second{0.793} & 0.772 & 0.708 & 0.742 & \third{0.791} & 0.752 & 0.750 & 0.599 & 0.791 & 0.780 & \best{0.793} & 0.727 \\

MNIST-C & ViT
& \best{0.739} & 0.721 & 0.682 & 0.723 & 0.705 & 0.697 & 0.718 & 0.543 & \third{0.732} & \second{0.733} & 0.723 & 0.566 \\

MVTec-AD & ResNet18
& \second{0.799} & 0.760 & 0.730 & 0.746 & 0.776 & \best{0.801} & 0.747 & 0.757 & 0.783 & \third{0.791} & 0.790 & 0.733 \\

MVTec-AD & ViT
& \best{0.896} & 0.813 & 0.753 & 0.815 & 0.866 & \third{0.880} & 0.799 & 0.841 & 0.718 & 0.864 & \second{0.891} & 0.672 \\

SVHN & ResNet18
& 0.657 & 0.583 & 0.585 & 0.583 & 0.613 & \second{0.665} & 0.592 & 0.637 & \best{0.673} & 0.654 & \third{0.659} & 0.597 \\

SVHN & ViT
& 0.579 & 0.514 & 0.515 & 0.531 & 0.563 & \best{0.594} & 0.529 & \second{0.584} & \third{0.584} & 0.550 & 0.580 & 0.526 \\

20News & BERT
& \best{0.626} & 0.546 & 0.555 & 0.560 & 0.592 & 0.572 & 0.561 & \second{0.617} & \third{0.606} & 0.597 & 0.603 & 0.567 \\

20News & RoBERTa
& \best{0.749} & 0.627 & 0.597 & 0.651 & 0.711 & \second{0.733} & 0.639 & 0.708 & \third{0.731} & 0.709 & 0.716 & 0.539 \\

Agnews & BERT
& \second{0.676} & 0.592 & 0.569 & 0.590 & \third{0.666} & 0.613 & 0.568 & \best{0.711} & 0.653 & 0.649 & 0.650 & 0.604 \\

Agnews & RoBERTa
& \best{0.881} & 0.701 & 0.616 & 0.690 & 0.832 & \third{0.833} & 0.674 & 0.659 & \second{0.865} & 0.804 & 0.813 & 0.523 \\

Amazon & BERT
& \third{0.608} & 0.578 & 0.548 & 0.564 & \second{0.608} & 0.594 & 0.571 & 0.573 & \best{0.608} & 0.606 & 0.604 & 0.525 \\

Amazon & RoBERTa
& 0.544 & 0.474 & 0.502 & 0.511 & 0.537 & \best{0.555} & 0.507 & \second{0.552} & 0.544 & 0.530 & 0.530 & \third{0.547} \\

Imdb & BERT
& 0.502 & 0.486 & 0.492 & 0.488 & 0.495 & \best{0.518} & 0.497 & 0.504 & \third{0.511} & 0.510 & \second{0.515} & 0.470 \\

Imdb & RoBERTa
& 0.530 & 0.432 & 0.493 & 0.487 & 0.519 & \second{0.540} & 0.494 & \best{0.543} & \third{0.537} & 0.511 & 0.504 & 0.506 \\

Yelp & BERT
& \second{0.679} & 0.628 & 0.579 & 0.605 & \best{0.684} & 0.672 & 0.603 & 0.656 & 0.673 & 0.660 & \third{0.676} & 0.529 \\

Yelp & RoBERTa
& 0.566 & 0.507 & 0.501 & 0.543 & \second{0.579} & \best{0.598} & 0.546 & \third{0.577} & 0.560 & 0.573 & 0.575 & 0.544 \\

\midrule
Average Rank
& & \best{2.60} & 8.60 & 10.65 & 8.80 & 5.25 & 4.25 & 8.50 & 6.70 & \third{4.20} & 4.60 & \second{3.95} & 9.90 \\
p-Value vs VAS
& & - & \smallscript{5.46e-20} & \smallscript{3.10e-25} & \smallscript{2.28e-23} & \smallscript{8.92e-18} & \smallscript{8.91e-10} & \smallscript{6.01e-23} & \smallscript{1.60e-14} & \smallscript{6.22e-11} & \smallscript{3.14e-14} & \smallscript{1.34e-08} & \smallscript{9.41e-24} \\
\bottomrule
\end{tabular}
\caption{AUROC across dataset families and encoders using each method's best hyperparameter setting. For multi-split datasets, we report average AUROC. \best{Orange} marks best, \second{blue} marks 2nd. Bottom-row p-values are paired Wilcoxon tests: VAS vs. each baseline over all 148 dataset-split/encoder instances.}
\label{tab:auroc_sweep}
\end{table*}
\begin{table*}[t]
\centering
\scriptsize
\setlength{\tabcolsep}{2.2pt}
\begin{tabular}{llcccccccccccc}
\toprule
Dataset & Encoder & VAS & ADERH & DIF & IForest & LUNAR & DN2 & LODA & LOF & KPCA & SLAD & GMD & DTE \\
\midrule

CIFAR10 & ResNet18
& \second{0.153} & 0.094 & 0.088 & 0.091 & 0.105 & \best{0.160} & 0.094 & 0.116 & 0.117 & \third{0.143} & 0.141 & 0.093 \\

CIFAR10 & ViT
& \second{0.659} & \third{0.544} & 0.221 & 0.405 & 0.441 & 0.341 & 0.427 & 0.054 & 0.075 & \best{0.689} & 0.491 & 0.108 \\

FashionMNIST & ResNet18
& \second{0.449} & 0.372 & 0.315 & 0.329 & 0.327 & 0.310 & 0.359 & 0.102 & 0.383 & \best{0.459} & \third{0.436} & 0.264 \\

FashionMNIST & ViT
& \best{0.546} & 0.510 & 0.205 & 0.433 & 0.329 & 0.317 & 0.435 & 0.073 & 0.393 & \second{0.537} & \third{0.511} & 0.129 \\

MNIST-C & ResNet18
& \second{0.237} & \third{0.237} & 0.169 & 0.183 & 0.196 & 0.190 & 0.226 & 0.109 & 0.220 & \best{0.255} & 0.235 & 0.169 \\

MNIST-C & ViT
& 0.226 & \best{0.273} & 0.167 & \second{0.243} & 0.195 & 0.209 & 0.235 & 0.082 & 0.219 & \third{0.238} & 0.201 & 0.100 \\

MVTec-AD & ResNet18
& \second{0.642} & 0.584 & 0.539 & 0.567 & 0.588 & \best{0.653} & 0.572 & 0.576 & 0.625 & 0.635 & \third{0.637} & 0.569 \\

MVTec-AD & ViT
& \second{0.748} & 0.652 & 0.554 & 0.635 & 0.722 & \third{0.734} & 0.613 & 0.687 & 0.554 & 0.692 & \best{0.750} & 0.468 \\

SVHN & ResNet18
& 0.092 & 0.077 & 0.071 & 0.074 & 0.081 & \second{0.096} & 0.077 & 0.084 & \best{0.098} & 0.092 & \third{0.093} & 0.077 \\

SVHN & ViT
& 0.066 & 0.055 & 0.054 & 0.056 & 0.062 & \second{0.068} & 0.056 & \best{0.069} & \third{0.067} & 0.060 & 0.066 & 0.056 \\

20News & BERT
& \second{0.088} & 0.067 & 0.067 & 0.066 & 0.080 & 0.074 & 0.065 & \best{0.094} & 0.082 & 0.077 & \third{0.084} & 0.072 \\

20News & RoBERTa
& \best{0.148} & 0.079 & 0.073 & 0.079 & 0.117 & \third{0.128} & 0.075 & 0.105 & \second{0.135} & 0.108 & 0.115 & 0.064 \\

Agnews & BERT
& \second{0.094} & 0.067 & 0.064 & 0.065 & \third{0.093} & 0.078 & 0.062 & \best{0.120} & 0.083 & 0.078 & 0.087 & 0.077 \\

Agnews & RoBERTa
& \best{0.334} & 0.122 & 0.076 & 0.092 & \third{0.232} & 0.228 & 0.087 & 0.096 & \second{0.292} & 0.183 & 0.213 & 0.057 \\

Amazon & BERT
& 0.063 & 0.060 & 0.057 & 0.058 & \second{0.063} & 0.061 & 0.059 & 0.058 & \best{0.063} & \third{0.063} & 0.062 & 0.054 \\

Amazon & RoBERTa
& 0.055 & 0.047 & 0.050 & 0.050 & 0.053 & \third{0.056} & 0.050 & \second{0.057} & 0.055 & 0.052 & 0.052 & \best{0.058} \\

Imdb & BERT
& 0.047 & 0.047 & 0.047 & 0.047 & 0.047 & 0.049 & 0.047 & \best{0.049} & 0.048 & \second{0.049} & \third{0.049} & 0.046 \\

Imdb & RoBERTa
& \third{0.056} & 0.045 & 0.050 & 0.051 & 0.053 & 0.054 & 0.051 & \best{0.057} & \second{0.057} & 0.053 & 0.052 & 0.051 \\

Yelp & BERT
& \second{0.086} & 0.075 & 0.067 & 0.070 & \best{0.090} & 0.083 & 0.068 & 0.082 & \third{0.084} & 0.079 & 0.083 & 0.057 \\

Yelp & RoBERTa
& 0.057 & 0.049 & 0.049 & 0.052 & \third{0.061} & \best{0.064} & 0.052 & \second{0.062} & 0.057 & 0.058 & 0.058 & 0.058 \\

\midrule
Average Rank
& & \best{3.05} & 7.60 & 10.60 & 8.75 & 5.40 & 4.65 & 8.40 & 6.30 & 4.70 & \third{4.35} & \second{4.30} & 9.90 \\
p-Value vs VAS
& & - & \smallscript{2.80e-14} & \smallscript{8.96e-26} & \smallscript{8.92e-20} & \smallscript{1.35e-17} & \smallscript{1.71e-08} & \smallscript{3.89e-18} & \smallscript{9.30e-13} & \smallscript{1.95e-13} & \smallscript{1.68e-05} & \smallscript{6.62e-11} & \smallscript{3.96e-23} \\
\bottomrule
\end{tabular}
\caption{AP across dataset families and encoders using each method's best hyperparameter setting. For multi-split datasets, we report average AP. \best{Orange} marks best, \second{blue} marks 2nd. Bottom-row p-values are paired Wilcoxon tests: VAS vs. each baseline over all 148 dataset-split/encoder instances.}
\label{tab:ap_sweep}
\end{table*}

\section{Alternative Heuristics for VAS}
\label{sec:choice_of_q}
We evaluate the sensitivity of VAS to alternative
data-dependent heuristics for selecting the Vendi Score
order $q$. Our default rule is
$q=\max\{2(1-\lambda_1),0.1\}$, where $\lambda_1$
is the largest eigenvalue of the trace-normalized
feature covariance matrix $\rho$.

We consider three families of alternatives:
\begin{align}
q_{\mathrm{slope}(a)}
&= \max\{a(1-\lambda_1),0.1\},
&& a\in\{1,5\},\\
q_{\mathrm{top}(k)}
&= \max\left\{2\left(1-\sum_{j=1}^k
\lambda_j\right),0.1\right\},
&& k\in\{5,10\},\\
q_{\mathrm{VS}_1}
&= \max\left\{
\frac{2\mathrm{VS}_1(\rho)}{D},0.1\right\}.
\end{align}
The slope variants modify the scaling factor in our
default rule, while the top-$k$ variants use the
cumulative mass of the leading eigenvalues.
The normalized Vendi Score heuristic instead uses
the effective spectral dimension, where
$\mathrm{VS}_1(\rho)=
\exp(-\sum_j\lambda_j\log\lambda_j)$.
We retain the lower bound of $0.1$ for numerical stability.

Table~\ref{tab:q_sensitivity} shows that our default
heuristic is robust to several reasonable changes in
the adaptation rule. Modifying the scaling factor from
2 to 1, or using the cumulative mass of the top 5 or
10 eigenvalues, produces similar AUROC values, with
no statistically significant differences under the
paired Wilcoxon tests. In contrast, increasing the
scaling factor to 5 or using the normalized
$\mathrm{VS}_1$ heuristic yields significantly worse
performance. These results support the effectiveness
of our simple default rule while indicating that VAS
does not depend strongly on its precise formulation.

\begin{table}[t]
\centering
\small
\setlength{\tabcolsep}{5pt}
\renewcommand{\arraystretch}{1.15}
\resizebox{\linewidth}{!}{%
\begin{tabular}{llcccccc}
\toprule
\textbf{Dataset} & \textbf{Encoder}
& \textbf{Adaptive}
& \textbf{Slope-1}
& \textbf{Slope-5}
& $\mathbf{VS_1/D}$
& \textbf{Top-5}
& \textbf{Top-10} \\
\midrule
CIFAR-10 & ResNet18 & 0.745 & 0.732 & 0.725 & 0.711 & 0.737 & 0.729 \\
CIFAR-10 & ViT & 0.941 & 0.914 & 0.943 & 0.894 & 0.919 & 0.911 \\
FashionMNIST & ResNet18 & 0.911 & 0.908 & 0.876 & 0.897 & 0.907 & 0.903 \\
FashionMNIST & ViT & 0.930 & 0.928 & 0.910 & 0.926 & 0.929 & 0.928 \\
MNIST-C & ResNet18 & 0.793 & 0.794 & 0.755 & 0.791 & 0.794 & 0.793 \\
MNIST-C & ViT & 0.739 & 0.741 & 0.737 & 0.743 & 0.740 & 0.741 \\
MVTec-AD & ResNet18 & 0.799 & 0.800 & 0.794 & 0.790 & 0.800 & 0.800 \\
MVTec-AD & ViT & 0.896 & 0.900 & 0.873 & 0.904 & 0.900 & 0.900 \\
SVHN & ResNet18 & 0.657 & 0.656 & 0.633 & 0.646 & 0.655 & 0.652 \\
SVHN & ViT & 0.579 & 0.585 & 0.546 & 0.589 & 0.585 & 0.586 \\
\midrule
20News & BERT & 0.626 & 0.629 & 0.613 & 0.630 & 0.629 & 0.629 \\
20News & RoBERTa & 0.749 & 0.749 & 0.749 & 0.750 & 0.749 & 0.749 \\
AG News & BERT & 0.676 & 0.678 & 0.663 & 0.680 & 0.678 & 0.678 \\
AG News & RoBERTa & 0.881 & 0.881 & 0.881 & 0.881 & 0.881 & 0.881 \\
Amazon & BERT & 0.608 & 0.608 & 0.607 & 0.608 & 0.608 & 0.608 \\
Amazon & RoBERTa & 0.544 & 0.544 & 0.544 & 0.544 & 0.544 & 0.544 \\
IMDB & BERT & 0.502 & 0.501 & 0.507 & 0.499 & 0.501 & 0.501 \\
IMDB & RoBERTa & 0.530 & 0.530 & 0.530 & 0.529 & 0.530 & 0.530 \\
Yelp & BERT & 0.679 & 0.679 & 0.678 & 0.679 & 0.679 & 0.679 \\
Yelp & RoBERTa & 0.566 & 0.566 & 0.566 & 0.566 & 0.566 & 0.566 \\
\midrule
\multicolumn{2}{l}{\textbf{Wilcoxon $p$-value}}
& -- & 0.399 & $1.46\times10^{-13}$
& 0.0058 & 0.371 & 0.0728 \\
\bottomrule
\end{tabular}%
}
\caption{
Sensitivity of VAS to alternative adaptive rules for
selecting $q$ across 148 ADBENCH benchmark instances.
AUROC values are averaged across dataset splits for
each dataset--encoder combination.
Mean AUROC is computed across all benchmark instances.
The bottom row reports unadjusted, one-sided paired
Wilcoxon tests comparing our default adaptive rule
against each alternative, with the alternative
hypothesis that our default achieves higher AUROC.
}
\label{tab:q_sensitivity}
\end{table}

\section{Anomaly Detection Across Different Anomaly Types}
\label{sec:anomaly_types}
\cite{han2022adbench} define four types of anomalies: local, global, dependency, and clustered anomalies. Our ImageNet experiments investigate clustered anomalies: coherent anomalous
subpopulations. Here we evaluate VAS on three additional constructions inspired by the remaining anomaly types. We create realistic anomalies using data from ADBENCH embedding datasets to construct
anomalies with different geometric properties. 

We evaluate four datasets (CIFAR-10, SVHN, 20News, and AG News), each with two embedding architectures, and compare VAS against five representative baselines: GMD, DN2, KPCA, LOF, and SLAD. Across all tables, we report mean AUROC over five independent trials and the average rank of each method across the eight dataset--encoder combinations.

\paragraph{Local Anomalies.}
We first investigate local anomalies - those that deviate from their local neighborhood. To construct these, we identify samples that are semantically different from a normal class but appear plausible under its local distribution. For each dataset, we split embeddings into fit and hold-out sets, fit one Gaussian per class, and assign each held-out sample to its highest-likelihood class.

For each predicted class $c$, we form a dataset from samples assigned to $c$. True-$c$ samples are normal; misclassified samples are anomalous. These are local anomalies because they lie in $c$'s region rather than being globally isolated. 

\begin{table}[t]
\centering
\scriptsize
\setlength{\tabcolsep}{5pt}
\renewcommand{\arraystretch}{1.15}
\resizebox{0.7\linewidth}{!}{%
\begin{tabular}{llcccccc}
\toprule
\textbf{Dataset} & \textbf{Encoder}
& \textbf{VAS}
& \textbf{GMD}
& \textbf{DN2}
& \textbf{KPCA}
& \textbf{LOF}
& \textbf{SLAD} \\
\midrule
CIFAR-10 & ResNet18
& 0.557 & 0.561 & 0.571 & 0.574 & 0.544 & 0.551 \\
CIFAR-10 & ViT
& 0.733 & 0.723 & 0.719 & 0.721 & 0.659 & 0.727 \\
SVHN & ResNet18
& 0.610 & 0.610 & 0.617 & 0.617 & 0.579 & 0.600 \\
SVHN & ViT
& 0.552 & 0.543 & 0.535 & 0.543 & 0.539 & 0.543 \\
\midrule
20News & BERT
& 0.555 & 0.560 & 0.526 & 0.525 & 0.525 & 0.548 \\
20News & RoBERTa
& 0.604 & 0.560 & 0.560 & 0.561 & 0.536 & 0.538 \\
AG News & BERT
& 0.661 & 0.608 & 0.609 & 0.609 & 0.620 & 0.632 \\
AG News & RoBERTa
& 0.660 & 0.589 & 0.642 & 0.630 & 0.624 & 0.605 \\
\midrule
\multicolumn{2}{l}{\textbf{Average Rank}}
& \textbf{1.81} & 3.63 & 3.56 & 3.06 & 5.19 & 3.75 \\
\bottomrule
\end{tabular}%
}
\caption{
AUROC for local class-confusable anomalies constructed
from genuine ADBENCH embeddings. Results are averaged
across assigned-class subsets and five independent trials.
Anomaly rates vary with class overlap.
Average ranks are computed across dataset--encoder
combinations (lower is better).
}
\label{tab:local_anomalies}
\end{table}

\paragraph{Global Directional Anomalies.}
We next investigate anomalies that deviate from the
global geometry of the normal data. We compute
class-centroid directions from the genuine embeddings
and define the subspace spanned by these centroids.
For each anomaly, we sample a random Gaussian
direction $z_\perp$ and project it onto the orthogonal
complement of this subspace. Given a normalized
genuine embedding $t$, we construct an anomalous
direction as
\begin{equation}
x_{\mathrm{anom}} =
\operatorname{normalize}
\left((1-\alpha)t+\alpha z_\perp\right),
\end{equation}
and restore the original embedding norm.

The parameter $\alpha$ controls the magnitude of
the directional perturbation. We report results
for $\alpha=0.1$, representing a relatively small
departure from the original embedding direction.
Each trial contains 2,500 genuine normal observations
and 278 synthesized anomalies, corresponding to
approximately 10\% contamination. Results are
averaged across five independent trials.

\begin{table}[t]
\centering
\scriptsize
\setlength{\tabcolsep}{5pt}
\renewcommand{\arraystretch}{1.15}
\resizebox{0.8\linewidth}{!}{%
\begin{tabular}{llcccccc}
\toprule
\textbf{Dataset} & \textbf{Encoder}
& \textbf{VAS}
& \textbf{GMD}
& \textbf{DN2}
& \textbf{KPCA}
& \textbf{LOF}
& \textbf{SLAD} \\
\midrule
CIFAR-10 & ResNet18
& 0.578 & 0.561 & 0.188 & 0.433 & 0.517 & 0.530 \\
CIFAR-10 & ViT
& 0.526 & 0.531 & 0.194 & 0.215 & 0.521 & 0.488 \\
SVHN & ResNet18
& 0.669 & 0.558 & 0.289 & 0.554 & 0.530 & 0.560 \\
SVHN & ViT
& 0.956 & 0.585 & 0.272 & 0.607 & 0.572 & 0.586 \\
\midrule
20News & BERT
& 0.996 & 0.719 & 0.727 & 0.746 & 0.919 & 0.924 \\
20News & RoBERTa
& 1.000 & 0.995 & 1.000 & 1.000 & 0.956 & 1.000 \\
AG News & BERT
& 0.991 & 0.712 & 0.657 & 0.752 & 0.932 & 0.896 \\
AG News & RoBERTa
& 1.000 & 1.000 & 1.000 & 1.000 & 1.000 & 1.000 \\
\midrule
\multicolumn{2}{l}{\textbf{Average Rank}}
& \textbf{1.63} & 3.69 & 5.13 & 3.75 & 3.94 & 2.88 \\
\bottomrule
\end{tabular}%
}
\caption{
AUROC for global directional anomalies constructed
by perturbing embeddings outside the
span of normal-class centroids.
Results use $\alpha=0.1$ and approximately 10\%
contamination, averaged across five trials.
Average ranks are computed across dataset--encoder
combinations (lower is better).
}
\label{tab:global_anomalies}
\end{table}

\paragraph{Feature-Dependency Anomalies.}
Finally, we investigate anomalies that preserve feature marginals while breaking feature dependence.. For each semantic class, we sample up to 2,500 genuine embeddings, normalize them to unit length, and apply a random orthogonal rotation. We divide the rotated features into four approximately equal-sized blocks.

To construct an anomaly, we replace one randomly selected block with the corresponding block from
a different observation. We then apply the inverse rotation and normalize the resulting embedding. This construction combines feature blocks from normal observations while disrupting their joint structure. We generate approximately 10\% anomalies and
average results across all available semantic classes and five random trials. For classes with fewer
observations, sample counts are reduced accordingly.

\begin{table}[t]
\centering
\scriptsize
\setlength{\tabcolsep}{5pt}
\renewcommand{\arraystretch}{1.15}
\resizebox{0.8\linewidth}{!}{%
\begin{tabular}{llcccccc}
\toprule
\textbf{Dataset} & \textbf{Encoder}
& \textbf{VAS}
& \textbf{GMD}
& \textbf{DN2}
& \textbf{KPCA}
& \textbf{LOF}
& \textbf{SLAD} \\
\midrule
CIFAR-10 & ResNet18
& 0.825 & 0.789 & 0.844 & 0.864 & 0.894 & 0.747 \\
CIFAR-10 & ViT
& 0.828 & 0.748 & 0.793 & 0.825 & 0.847 & 0.669 \\
SVHN & ResNet18
& 0.858 & 0.778 & 0.851 & 0.841 & 0.879 & 0.791 \\
SVHN & ViT
& 0.959 & 0.694 & 0.799 & 0.818 & 0.835 & 0.716 \\
\midrule
20News & BERT
& 0.922 & 0.685 & 0.814 & 0.744 & 0.895 & 0.775 \\
20News & RoBERTa
& 0.985 & 0.549 & 0.728 & 0.779 & 0.762 & 0.628 \\
AG News & BERT
& 0.917 & 0.664 & 0.754 & 0.736 & 0.910 & 0.770 \\
AG News & RoBERTa
& 0.999 & 0.531 & 0.810 & 0.814 & 0.827 & 0.611 \\
\midrule
\multicolumn{2}{l}{\textbf{Average Rank}}
& \textbf{1.63} & 5.75 & 3.63 & 3.38 & 1.75 & 4.88 \\
\bottomrule
\end{tabular}%
}
\caption{
AUROC for feature-dependency anomalies constructed
by swapping blocks of orthogonally rotated embeddings.
Results use approximately 10\% contamination
and are averaged across semantic classes and five trials.
Average ranks are computed across dataset--encoder
combinations (lower is better).
}
\label{tab:dependency_anomalies}
\end{table}

Overall, VAS achieves the best average rank across
all three anomaly constructions. It performs particularly
well on global directional anomalies and on
feature-dependency anomalies in text embeddings.
For image-based dependency anomalies, LOF is also
highly competitive. Local class-confusable anomalies
are substantially more challenging across methods,
although VAS retains the strongest average rank.
These experiments provide complementary evidence
of VAS's ability to identify different forms of
anomalous structure in pretrained embedding spaces.

\end{document}